\documentclass{article}

\usepackage{makeidx}
\makeindex
\usepackage{amsmath, amssymb, amsthm, verbatim, url,varioref}
\usepackage{fullpage}
\usepackage{tikz}
\usetikzlibrary{arrows.meta, positioning, fit, backgrounds}
\newcommand{\E}                 {\Bbb{E}}
\newcommand{\KL}{\operatorname{KL}}
\newcommand{\reals}{\Bbb{R}}
\newcommand{\belief}{\pi}
\newcommand{\Belief}{\Pi(V)}
\newcommand{\emb}{{W}_{\text{emb}}}
\newcommand{\softmax}{\operatorname{softmax}}
\newcommand{\LayerNorm}{\operatorname{LayerNorm}}
\newcommand{\FFN}{\operatorname{FFN}}
\newcommand{\MultiHead}{\operatorname{MultiHead}}
\newcommand{\concat}{\operatorname{concat}}

\newcommand{\Embed}{E}

\newcommand{\unit}{\mathbf{1}}
\newcommand{\p}{\top}
\newcommand{\blockdiag}{\operatorname{blockdiag}}
\newcommand{\nhead}{n_{\text{head}}}
\newcommand{\Layer}{\mathsf{Layer}}
\newcommand{\Attn}{\operatorname{Attn}}
\newcommand{\defn}{\stackrel{\text{defn}}{=}}
\newcommand{\Simplex}{\Pi}
\newcommand{\AR}{f}
\newcommand{\I}{\mathcal{C}}
\newcommand{\CW}{\Delta}
\newcommand{\nanoGPT}{\texttt{nanoGPT}}
\newcommand{\code}[1]{\texttt{#1}}
\newcommand{\nanochat}{\texttt{nanochat}}

\usepackage{enumitem}
\setlist[itemize]{noitemsep,topsep=0pt,parsep=0pt,partopsep=0pt}
\setlist[enumerate]{noitemsep,topsep=0pt,parsep=0pt,partopsep=0pt}

\usepackage{listings}
\usepackage{xcolor}

\newcounter{mypart}
\newcommand{\partsection}[1]{%
  \stepcounter{mypart}%
  \section*{\Large\bfseries Part~\arabic{mypart}: #1}%
  \addcontentsline{toc}{section}{\protect\numberline{}Part~\arabic{mypart}: #1}%
}

\makeatletter
\@addtoreset{subsection}{mypart} % reset subsection when mypart increments
\makeatother

\usepackage[margin=1in]{geometry}
\begin{document}

\title{LLMs as High-Dimensional Nonlinear Autoregressive Models with Attention: Training, Alignment and Inference}

%\title{A Mathematical Framework for LLMs: Autoregressive Models with Attention and Alignment}

%\title{LLMs as High-Dimensional Nonlinear Autoregressive Models with Attention: Training, Alignment, and Inference}
\author{Vikram Krishnamurthy, ECE, Cornell University,  vikramk@cornell.edu} 
\maketitle

\begin{abstract}
Large language models (LLMs) based on transformer architectures are typically described through collections of architectural components and training procedures, obscuring their underlying computational structure.
This review article provides a concise mathematical reference for researchers seeking an explicit, equation-level description of LLM training, alignment, and generation. We formulate LLMs as high-dimensional nonlinear autoregressive models with attention-based dependencies.
The framework encompasses pretraining via next-token prediction, alignment methods such as reinforcement learning from human feedback (RLHF), direct preference optimization (DPO), rejection sampling fine-tuning (RSFT), and reinforcement learning from verifiable rewards (RLVR), as well as autoregressive generation during inference.
Self-attention emerges naturally as a repeated bilinear--softmax--linear composition, yielding highly expressive sequence models.
This formulation enables principled analysis of alignment-induced behaviors (including sycophancy), inference-time phenomena (such as hallucination, in-context learning, chain-of-thought prompting, and retrieval-augmented generation), and extensions like continual learning, while serving as a concise reference for interpretation and further theoretical development.
\end{abstract}

{\bf Keywords}. Large language models, transformers, nonlinear autoregressive models, self-attention,
LLM alignment, RLHF, DPO, RLVR, RSFT, hallucination, sycophancy, retrieval-augmented generation (RAG),
in-context learning, chain-of-thought

\tableofcontents
\clearpage

%The notes in this document are intentionally elementary. They were written so that I can better understand the internal workings of an LLM from a mathematical perspective during training, alignment, and generation.
%A careful mathematical model reveals  what the LLM computes, making its behavior analyzable rather than mysterious.
%Please email me if you have  suggestions or find any errors. I hope to teach this in my undergraduate class  ECE 2720 at Cornell.

\section*{Introduction}

Large Language Models (LLMs) based on transformer architectures have become central to modern machine learning systems. Their behavior, however, is often described through a collection of architectural components and training heuristics -- pretraining objectives, alignment procedures, attention mechanisms, and prompting strategies  -- rather than through a unified mathematical viewpoint that spans the full lifecycle of an LLM. This can obscure the relationship between training, alignment, and inference, and makes it difficult to reason analytically about model behavior.

This  paper  reviews and synthesizes a mathematically transparent formulation of LLMs  as a high-dimensional, high-order  nonlinear autoregressive model with attention.
Here, ``dimension'' refers to  the dimension of the latent hidden state produced by the transformer at each time step (typically $10^3-10^4$), while  the ``order'' refers to the number of past tokens on which the next-token distribution depends, i.e., the  temporal context window (typically  $\approx 10^5$).
The exposition is  elementary and emphasizes abstraction over architectural detail. Rather than introduce new algorithms or performance improvements, we  clarify what an LLM computes at each stage, and  express these computations in forms that are amenable to analysis.

The paper is organized conceptually as follows. Part 1 reviews pretraining, where an LLM is fit via next-token prediction and can be viewed as learning a high-dimensional conditional distribution over token sequences. Part 2  discusses alignment training, including reinforcement learning from human feedback (RLHF),  direct preference optimization (DPO), reinforcement learning with  verifiable rewards (RLVR) and rejection sampling fine tuning (RSFT), and explains how these procedures modify the pretrained LLM to shape its responses. Part 3 focuses on generation (inference), where the trained and aligned LLM is used to produce tokens autoregressively. In this part, we present a unified abstraction of LLM generation as a high-dimensional nonlinear autoregressive model with attention, which serves as a mathematical reference point for the remainder of the paper.

Building on this formulation, Part 4 considers extensions beyond static training, including continual learning under streaming data with selective parameter updates, as well as alignment-induced artifacts such as sycophancy (a static bias toward excessive agreement with user input), inference-time phenomena such as hallucination (dynamical drift during autoregressive rollout into fluent but unsupported continuations), and mitigation strategies such as retrieval-augmented generation (RAG) for grounding responses in external evidence. Finally, Part 5 examines prompting strategies during inference, highlighting how in-context learning (ICL) and chain-of-thought (CoT) prompting can be understood as emergent properties of the same autoregressive generation mechanism when conditioned on structured prompts.

%Building on this formulation, Part 4 considers extensions beyond static training, including continual learning under streaming data with selective parameter updates, as well as phenomena such as sycophancy, which can be interpreted as an artifact of alignment objectives interacting with the underlying autoregressive structure. Finally, Part 5 examines in-context learning (ICL) and chain-of-thought (CoT) prompting during inference, highlighting how these behaviors can be understood as emergent properties of the same autoregressive generation mechanism when conditioned on structured prompts.

\paragraph{ How does an  LLM generate tokens?}
As a short appetizer (elaborated later), we outline how
an LLM generates new tokens. An LLM can be viewed as a $\CW$-order
nonlinear autoregressive (AR) model, where $\CW$  denotes   \index{context window}
the context window length. After training, the optimized parameters $\theta^\star$ of the transformer neural  network are fixed.
The user-provided \textit{prompt} consists of the initial tokens
$x_1,\ldots,x_{t_0}$. For $t \ge t_0$, subsequent tokens
$x_{t+1}, x_{t+2}, \ldots$ are generated autoregressively by the LLM as follows:
At time $t$, the LLM uses the previous up to $\CW$ tokens\footnote{Indices corresponding to $t<1$ are omitted.}
$x_{t-\CW+1:t}$ to predict the next token. These tokens constitute the
model's \textit{context}.  \index{context}
The LLM then draws the next token $x_{t+1}$ randomly from the
$V$-dimensional probability vector $\belief_t^\tau$
(where $V$ denotes the size of the LLM vocabulary):
\begin{equation}\label{eq:AR_gen}
x_{t+1}
\sim
\belief_t^\tau
=
\softmax\!\left(
\frac{W^*_{\mathrm{out}}\,
      \AR_{\theta^\star,t}\!\bigl(x_{t-\CW+1:t}\bigr)
      + b^*_{\mathrm{out}}}{\tau}
\right),
\qquad \tau>0 .
\end{equation}
In the above AR model, 
$\tau$ is the user-chosen sampling temperature, and  $W^*_{\mathrm{out}}\in \mathbb{R}^{V \times d}$, $b^*_{\mathrm{out}}\in \mathbb{R}^V$ are learned  parameters, where $d$ denotes the latent hidden-state (embedding) dimension of the transformer. Also for any $z\in \reals^V$, $\softmax(z) $ is the $V$-dimensional probability vector with elements $\exp(z_i)/\sum_j \exp(z_j)$, $i=1\ldots, V$.

The autoregressive feature map $\AR_{\theta,t}\in \reals^d$ appearing in~\eqref{eq:AR_gen} is  the final-layer hidden state corresponding to the most
recent token $t$, defined by the composition of
 $L$ layers of the transformer:
\begin{equation}\label{eq:AR_comp}
\AR_{\theta,t}\!\bigl(x_{t-\CW+1:t}\bigr)
\;\triangleq\;
h_{L,t},
\qquad
h_{\ell,t}  = g^{(\ell)}_\theta\!\bigl(h_{\ell-1,t-\CW+1}, \ldots  ,h_{\ell-1,t} \bigr),
%=
%g^{(\ell)}_\theta\!\bigl(\{h_{\ell-1,s}\}_{s=t-\CW+1}^{t} \bigr),
\quad \ell=1,\ldots,L,
\end{equation}
initialized at layer 0 by
$ (h_{0,t-\CW+1}, \ldots, h_{0,t})  = \emb[x_{t-\CW+1:t}]$,
% $\{h_{0,s}\}_{s=t-\CW+1}^t := \emb[x_{t-\CW+1:t}]$.
where $\emb \in \reals^{d \times V}$ denotes the learned token embedding
matrix that maps each discrete token to $\reals^d$. 
We refer to  $h_{\ell,s} \in \reals^d$ as the hidden-state representation outputted at layer $\ell$ for token~$s$.  Detailed constructions are given in~\eqref{eq:hidden_state_matrices}, \eqref{eq:attention_scores}.

Let us point out 
the crucial feature of~\eqref{eq:AR_comp}, namely  the 
self-attention mechanism in each layer $\ell$: 
\begin{equation}\label{eq:g_layer}
g^{(\ell)}_\theta\!\bigl(h_{\ell-1,t-\CW+1},\ldots,h_{\ell-1,t}\bigr)
=
\phi\!\left(
\sum_{s=t-\CW+1}^{t}
\softmax_s\!\bigl(a^{(\ell)}_\theta(t)\bigr)\,
L^{(\ell)}_\theta\, h_{\ell-1,s}
\right),
\end{equation}
where $a^{(\ell)}_\theta(t)\in \reals^\CW$ denotes the vector
of attention scores with real-valued elements 
\[ a^{(\ell)}_\theta(t,r)
\;\defn\;
\big\langle M^{(\ell)}_\theta\, h_{\ell-1,t},\;
          N^{(\ell)}_\theta\, h_{\ell-1,r}\big\rangle,
\qquad r=t-\CW+1,\ldots,t, \]
and  $\softmax_s(z)$ denotes the $s$-th component of the $\CW$-dimensional 
probability vector $\softmax(z)$.
Also,  $M^{(\ell)}_\theta$, $N^{(\ell)}_\theta$ and $L^{(\ell)}_\theta$ are
learned matrices. Finally, the nonlinearity $\phi(\cdot)$ in~\eqref{eq:g_layer} abstracts the combined
effect of residual, normalization, and feedforward operations within a
transformer block.

%$\phi(\cdot)$ in~\eqref{eq:g_layer}  is a fixed pointwise nonlinearity induced by the transformer architecture; pointwise means $\phi$ acts independently on each coordinate of its vector argument. 

{\bf Summary}.
There are two main takeaways: First, tokens are drawn from probability vector~\eqref{eq:AR_gen} specified by a $\CW$-order AR model, where the dimension of the hidden state is $d$. Second,  inside the softmax operator lies the 
all-important  attention mechanism~\eqref{eq:g_layer} which is a bilinear function of the hidden
states $h$ inside the softmax, followed by a linear  (weighted-sum) aggregation   of $h$.
Remarkably,  this  bilinear–softmax–linear mechanism, when composed repeatedly
across  multiple layers of a transformer neural network, makes an LLM  appear almost  sentient, despite
the absence of explicit symbolic reasoning or  any predesigned task-specific user interface.

{\bf Context}. LLMs are the modern successors to Word2Vec — replacing static embeddings with deep, contextual representations learned end-to-end via autoregressive training on Transformers with self-attention.

Examples of closed source LLMs include   ChatGPT (GPT-5 era), Claude (Opus 4.5/Sonnet 4), Gemini 3 and Grok 4.1. Examples of 
open source LLMs include  Llama 4 (Scout/Maverick), DeepSeek-V3.2, Qwen3, Mistral Large 3 and  GLM-4.7.

\begin{figure}[h]
  \centering
    \scalebox{0.9}{
\begin{tikzpicture}[
  node distance=3.3cm,
  block/.style={rectangle, draw, rounded corners,
                minimum width=3.4cm, minimum height=1.2cm,
                align=center},
  line/.style={-Stealth, thick},
  dashedline/.style={-Stealth, thick, dashed}
]

% Main pipeline
\node[block] (pre) {Pretraining\\(Part 1)};
\node[block, right=of pre] (align) {Alignment Training\\(Part 2: RLHF / RLVR)};
\node[block, right=of align] (infer) {Inference (Generation) \\(Part 3)\\ Includes  ICL, CoT, RAG};

\draw[line] (pre) -- node[above, align=center]
{$\theta^{\text{pre}}$ \\ (pretrained}
 node[below, align=center]
  {parameters)}
(align);

\draw[line] (align) -- node[above, align=center]
{$\theta^{\text{aligned}}$ \\ (aligned}
node[below, align=center]
  {chat model)}
(infer);

% Continual learning box below inference
\node[block, below=1.8cm of infer]
  (cont) {Continual Learning \\ with new data \\(Part 4)};

% Feedback loop arrows
\draw[<->, >=latex,thick,dashed] (infer.south) -- node[right, align=center]
  {new data} (cont.north);

%\draw[dashedline] (cont.west) -| node[below, pos=0.3, align=center]
 % {updated $\theta_{\text{trainable}}$} (infer.south);

\end{tikzpicture}
}
\caption{Schematic overview of the paper structure}
  \label{fig:outline}
\end{figure}
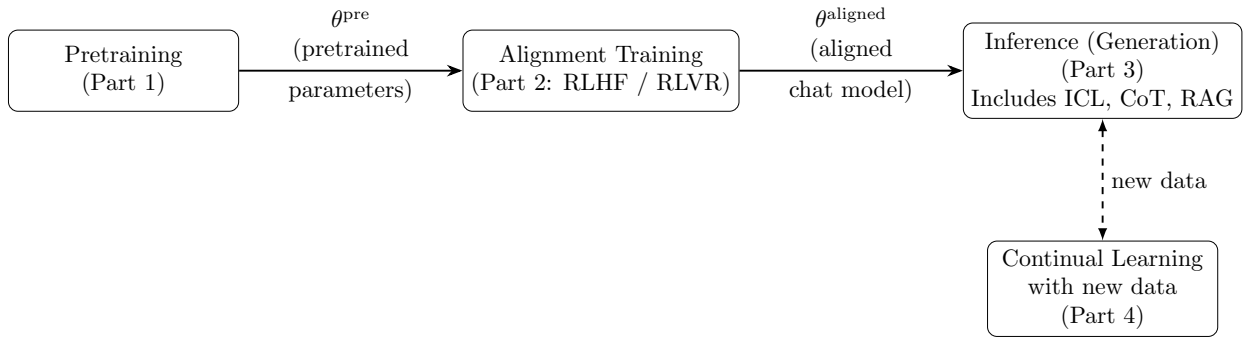

Training a modern LLM  has  two stages:
\begin{itemize}
  \item {\em Part 1. Pretraining},  where the LLM learns its core language ability by 
    predicting the next token on a large text corpus;
\item  \emph{Part 2. Alignment training}, 
where the pretrained LLM is fine-tuned using human feedback to make its 
behavior helpful, honest, and safe.
\end{itemize}
 Pre-training and Alignment Training function constitute the  pre-deployment foundation where an LLM  is first endowed with vast world knowledge and then  refined to follow instructions with logical and ethical precision. 
Once trained, the LLM is deployed and used for text generation (inference);  this is discussed in \textit{Part~3}.
Finally \textit{Part~4} discusses \textit{continual learning} (after deployment of an LLM) to incorporate new information  without eroding its foundational alignment or suffering from catastrophic forgetting, 
and \textit{sycophancy}, which is a known artifact of preference-based  alignment training.
Figure~\ref{fig:outline} shows this organization  schematically.

\paragraph{Notation and Code.}
\nanoGPT{}~\cite{karpathy_nanogpt}\footnote{The code snippets presented in this paper are   from the now-deprecated \nanoGPT{} repository~\cite{karpathy_nanogpt} (last major update 2024) due to its  pedagogical simplicity and minimalism, which clearly illustrates the core  concepts of autoregressive generation without additional implementation complexity.  Andrej Karpathy's successor project \nanochat{} (active development as of January 2026) provides an updated, full-stack implementation of pretraining, supervised fine-tuning, RLVR-style alignment, and efficient autoregressive generation with KV caching. See \url{https://github.com/karpathy/nanochat}.} is a minimal, open-source PyTorch implementation by Andrej Karpathy that faithfully reproduces the GPT-2 architecture (e.g., the 124M parameter model), enabling training and finetuning of medium-sized autoregressive language models from scratch.
The complete code for training and token-by-token generation is available in the \nanoGPT{} repository on GitHub (MIT License).\footnote{\url{https://github.com/karpathy/nanoGPT}}
For practical work with a wide range of pretrained transformer models (including GPT-2, LLaMA, Mistral, and many others), the Hugging Face \texttt{transformers} library~\cite{WDS20} has become the standard tool in the research and engineering community.
Note that coding implementations of LLMs typically use row vectors for efficiency\footnote{PyTorch and NumPy store tensors in row-major order, so treating token embeddings as row vectors (with the embedding dimension as the last index) ensures contiguous memory access during linear operations and attention computations, leading to better cache performance and higher computational efficiency.}, \textit{whereas the current paper  adopts the classical mathematical convention of column vectors throughout}. This distinction is essential for interpreting the equations in subsequent sections.

\partsection{Pre-Training an LLM}

\subsection{Vocabulary, Corpus and Token Embeddings} 

\paragraph{Vocabulary}
The LLM operates over a fixed vocabulary
\[
\mathcal{V} = \{ w_1, w_2, \dots, w_V \},
\]
where the vocabulary size  $V$ is typically between $50,000$ to $100,000$  Byte Pair Encoding (BPE) tokens.  \index{tokens!BPE} \index{BPE (Byte Pair Encoding)}
A BPE token is generally smaller than  a word and larger than a character: examples include  common prefixes
{\tt re, pre}, root words {\tt run, talk, happi},  common function words {\tt the, of, in},
punctuation {\tt  ,   ? .  ! }, common words  {\tt democracy, integral }.

Because BPE tokenization is learned in a data-driven manner from a training corpus,
it can sometimes produce counter-intuitive segmentations. For example, the phrase
“solid gold” may be tokenized as two tokens {\tt solid} and {\tt\ gold} (preserving
their individual meanings), whereas a rarer compound such as “solidgold” may remain
a single token or be split differently, depending on corpus statistics.
\footnote{BPE tokenization is the primary reason why LLMs exhibit specific cognitive gaps: 
\begin{itemize}[noitemsep,topsep=0pt]
    \item \textbf{Arithmetic}: The sequence ``123'' may be mapped to a single index $i \in \mathcal{V}$, while ``124'' may be split into two indices $(j, k)$. This inconsistency in token boundaries prevents the model from learning standard digit-based carryover rules.
    \item \textbf{Spelling}: Since the model processes the token for ``apple'' as a single atomic vector $e_t$, it lacks a direct representation of the character sequence $\{a, p, p, l, e\}$. Character-level awareness only emerges if the training corpus contains explicit spelling tasks (e.g., ``a-p-p-l-e'').
\end{itemize}}

\paragraph{Corpus of Training Sequences}
The training corpus for an LLM is a collection of token sequences where each token\footnote{We identify each token in the vocabulary $ \mathcal{V} = \{ w_1, w_2, \dots, w_V \}$ with its integer index in $\{1,\ldots, V\}$. Thus, $x_t^{(n)}$ denotes the token index, and we freely refer to it as the corresponding token when no confusion arises.}
  $x^{(n)}_t \in \mathcal{V}$.
\[
\mathcal{D}
=
\Big\{
(x^{(n)}_1, \dots, x^{(n)}_{T_n})
\;\big|\;
n=1,\dots,N
\Big\},
\]
with $T_n$ the length of document $n$.  Typically $T_n \approx 16$k (per document).
A small fraction of documents are long (entire books).
The LLM is trained on  $N$ documents (sequences) such as news articles, Wikipedia pages, blog posts, etc.
LLMs train on  total number of tokens $\sum_{n=1}^N T_n$ of $1$–$20$ trillion tokens.

\paragraph{Token Embeddings}
The LLM  learns an \textit{embedding matrix} 
\[
\emb \in \reals^{d  \times V},
\]
and each token is mapped to a $d$-dimensional column vector (where $d$ is an even integer) \index{embedding!token}
\[
e^{(n)}_t = \emb[x^{(n)}_t] \in \mathbb{R}^d,
\qquad
d = 4096 \text{ for Llama-3 and } d = 12288 \text{ for GPT-4 (estimated)}
\]
i.e., $e_t^{(n)}$ is the $x_t^{(n)} $-th column of matrix $\emb$.
%(The column index is the \emph{token value} $x_t^{(n)}$, not the position~$t$.)
These embeddings form the input to the transformer.

\subsection{Causal Transformer Architecture as a Nonlinear Map}

Stack the embeddings of the  sequence of $T_n$ tokens in  document $n$ into the $d \times T_n$  matrix
\[
\Embed^{(n)} \defn
\begin{bmatrix} 
e^{(n)}_1 & 
e^{(n)}_2 & 
\cdots 
& e^{(n)}_{T_n} 
\end{bmatrix}  = \emb[x^{(n)}_{1:T_n}]
\in \reals^{d\times T_n}.
\]

\paragraph{Hidden Latent State} 
A transformer neural network with model parameters $\theta$  is a deep nonlinear map
$$F_\theta : \mathbb{R}^{d\times T_n} \to \mathbb{R}^{d \times T_n}$$
composed of multiple layers of masked self-attention and feedforward networks.
Applied to the entire sequence in parallel, it outputs the {\em hidden state} matrix
(internal, context-dependent representations of tokens) \index{hidden latent state}
\begin{equation}
  \label{eq:transformer_1}
S^{(n)} = F_\theta(\Embed^{(n)})
=
\begin{bmatrix}
{s^{(n)}_1} &
{s^{(n)}_2} &
\cdots &
{s^{(n)}_{T_n}} 
\end{bmatrix}  \in \reals^{d \times T_n}
\end{equation}
\textit{Causal masking} ensures that the  hidden state $s^{(n)}_t \in \reals^d$ for token $t$ outputted by the transformer depends only on recent tokens in the context window, namely, the  embeddings  $e^{(n)}_{t-\CW+1},\dots,e^{(n)}_t$,
or equivalently on the tokens $x^{(n)}_{t-\CW + 1}, \dots, x^{(n)}_{t}$. Typically the context window length $\CW \approx 10^5$.  \index{context window}
Let  $\unit_t$ denote the unit vector in $\reals^{T_n}$ with 1 in the $t$-th position. Note that  the 
hidden state   $s^{(n)}_t $ is 
\begin{equation}
  s^{(n)}_t =  F_\theta(\Embed^{(n)}) \,\unit_t = F_\theta(\emb[x_{1:T_n}^{(n)} ]) \,\unit_t \in \reals^{d}.  \label{eq:causal_masking}
\end{equation}
The  causal  masking constraint  on the transformer neural network mapping $F_\theta$ imposes the constraint \index{causal masking constraint} 
$$\frac{ \partial  s^{(n)}_t}{\partial e^{(n)}_\tau} = 0 \quad  \tau \notin  \{t-\CW+1,\dots,t\}   \quad \text{ which implies } 
\frac{ \partial F_\theta(\Embed^{(n)}) }{\partial e^{(n)}_\tau} \,\unit_t
= 0, \quad  \tau \notin  \{t-\CW+1,\dots,t\}. $$
That is, causal masking imposes a  banded triangular structural  constraint on the Jacobian of $F_\theta$.
Later in this document, \eqref{eq:hidden_state_matrices} and \eqref{eq:attention_scores}  give  precise expressions of this causality within the self attention mechanism.

{\em Remark}. We can write the $\CW$-order AR structure of the hidden state more explicitly as
\begin{equation}
  s^{(n)}_t
  \;\defn\; \AR_{\theta,t}\bigl( x^{(n)}_{t-\CW+1:t}\bigr) \in\; \reals^d , \quad
\text{ where } \AR_{\theta,t}(\cdot) 
\defn  \bigl(F_\theta(\emb[\cdot])\bigr)\,\unit_t ,
    \label{eq:causal_masking2}
  \end{equation}
and  indices
corresponding to $t<1$ are omitted.
%where we define the induced causal map  $\AR_{\theta,t}(\cdot) 
%\defn  \bigl(F_\theta(\emb[\cdot])\bigr)\,\unit_t $.

\paragraph{Output Layer: Logits} 
Each hidden state $ s^{(n)}_t \in \reals^d$ outputted by the transformer is mapped to a vector of next-token logits:
\begin{equation}
  \label{eq:next_logit}
z^{(n)}_t = W_{\text{out}}\, s^{(n)}_t + b_{\text{out}}
\in \mathbb{R}^{V},
\end{equation}
with trainable parameters $W_{\text{out}} \in \mathbb{R}^{V \times d}$ and
$b_{\text{out}} \in \mathbb{R}^V$.

The precise construction of the transformer mapping $F_\theta$, including multi-head causal self-attention and modern Rotary Position Embeddings (RoPE),  is developed in Appendix~\vref{app:rope}.

\subsection{Pre-Training Objective}

\paragraph{Softmax Prediction}
Given the logit vector $z^{(n)}_t \in \reals^V$, the next-token distribution (predicted distribution of next token) is    \index{softmax  prediction}
\begin{equation}
  \label{eq:softmax_predict}
  \hat{\belief}^{(n)}_t \defn \begin{bmatrix} \Pr_\theta\!\big(x^{(n)}_{t+1}=w_1 \mid x^{(n)}_{1:t}\big) \\
    \vdots \\ \Pr_\theta\!\big(x^{(n)}_{t+1}=w_V \mid x^{(n)}_{1:t}\big)
    \end{bmatrix} = 
\softmax(z^{(n)}_t) \defn \begin{bmatrix}
  \frac{\exp(z^{(n)}_{t,1})}
  {\sum_{j=1}^{V} \exp(z^{(n)}_{t,j})} \\ \vdots \\
   \frac{\exp(z^{(n)}_{t,V})}
  {\sum_{j=1}^{V} \exp(z^{(n)}_{t,j})}
\end{bmatrix}  \in \Belief.
\end{equation}
Here  $\Belief$ denotes  the space of $V$ dimensional probability vectors. Recall $\softmax(\cdot)$ maps a vector in Euclidean space to a probability vector (vector in unit simplex) of the same dimension.

\paragraph{Minimizing Cross Entropy Loss}  \index{cross entropy loss}
The aim is to compute the model parameters (neural network weights) that minimize the cross entropy between the predicted distribution and the empirical distribution induced by the data corpus. The loss function is 
\[
\mathcal{L}(\theta)
=
\sum_{n=1}^N
\sum_{t=1}^{T_n - 1}
\KL\!\left(
\unit_{x_{t+1}^{(n)}},
\;\hat{\belief}_t^{(n)}
\right),
\qquad \text{ where } 
\unit_{x_{t+1}^{(n)}}(i)
= \mathbf{1}\{i = x_{t+1}^{(n)}\}.
\]
Note the Kullback Leibler  Divergence between two pmfs $\alpha$ and $\beta$ is
$$ \KL(\alpha,\beta) = \sum_{i=1}^V \alpha(i) \log \frac{\alpha(i)}{\beta(i)} $$
Since the target is a one-hot distribution, this simplifies to the negative log-likelihood:
\begin{equation}
  \label{eq:loss_entropy}
\mathcal{L}(\theta)  = -  \sum_{n=1}^N \sum_{t=1}^{T_n-1}  \sum_{i=1}^V
\mathbf{1}\{x^{(n)}_{t+1} = w_i\}
\log \hat \belief_t^{(n)}(w_i)= - \sum_{n=1}^N \sum_{t=1}^{T_n-1}  \log \hat{\belief}^{(n)}_t\!\big(x^{(n)}_{t+1}\big)
\end{equation}

%\subsection*{8. Training Problem}
The model parameters are learned by solving
\[
\boxed{
\theta^\star = \arg\min_{\theta} \mathcal{L}(\theta), \quad 
\theta = \big(\emb,\; \text{all transformer-layer weights},\; W_{\text{out}},\; b_{\text{out}}\big).}
\]
To train GPT-4 from scratch, OpenAI used a supercomputer cluster with 25,000 NVIDIA A100 GPUs running in parallel for 100 days. Total compute: $3 \times  10^{23}$  FLOPs.
Cost: ~\$100 million (mostly electricity, hardware, and cloud fees).

(vi) In \nanoGPT{}, the training objective is exactly this negative log-likelihood~\eqref{eq:loss_entropy}.
The final logits $z^{(n)}_t$ are produced by a linear layer (\texttt{lm\_head}) typically weight-tied with the token embeddings.
The cross-entropy loss is computed only when targets are supplied, using PyTorch's numerically stable implementation.
Key excerpt from the \texttt{GPT.forward} method in \texttt{model.py}:

\begin{lstlisting}[language=Python, caption={Loss computation~\eqref{eq:loss_entropy} in nanoGPT (from \texttt{model.py})}]
logits = self.lm_head(x)  # (B, T, vocab_size)

# if targets provided, compute cross-entropy loss
loss = None
if targets is not None:
    B, T, C = logits.shape
    logits = logits.view(B*T, C)
    targets = targets.view(B*T)
    loss = F.cross_entropy(logits, targets)

return logits, loss
\end{lstlisting}

During training (in \code{train.py}), the model is called with shifted targets to compute the loss over all positions in parallel:

\begin{lstlisting}[language=Python, caption={Training step excerpt from nanoGPT (from \code{train.py})}]
# forward pass
logits, loss = model(idx, targets)  # idx: input tokens, targets: idx shifted by 1

# backward pass
optimizer.zero_grad()
loss.backward()
optimizer.step()
\end{lstlisting}

This implements teacher-forcing autoregressive training, minimizing the exact negative log-likelihood summed over all next-token predictions in the batch.

{\em Remark.} 
The vector $z^{(n)}_t \in \reals^V$ from~\eqref{eq:next_logit} is called the {\em logits} in deep learning code.
The predicted distribution is $\hat{\belief}^{(n)}_t = \softmax(z^{(n)}_t)$~\eqref{eq:softmax_predict}.
PyTorch's \code{F.cross\_entropy} computes the loss directly from logits and integer targets $x_{t+1}^{(n)}$ without explicit softmax, yielding precisely $\mathcal{L}(\theta)$ in~\eqref{eq:loss_entropy}.

\subsection{Summary (Viewed as a Large Nonlinear Autoregressive Model)}

For each token position $t$ in document $n$, the LLM predicts the distribution of the next
token $x^{(n)}_{t+1}$ as  \index{nonlinear AR model}
\[
\hat{\belief}^{(n)}_t
=
\softmax\!\left(
W_{\text{out}}\, s^{(n)}_t + b_{\text{out}}
\right) \in \Belief, 
\qquad
s^{(n)}_t =  F_\theta(\Embed^{(n)}) \,\unit_t = F_\theta(\emb[x_{1:t}^{(n)}]) \,\unit_t \in \reals^{d}.
\]
Notice that  prediction  of $x^{(n)}_{t+1}$  uses the entire prefix $x^{(n)}_{1:t}$. Hence the above representation is a large
($\CW$-order)  nonlinear   autoregressive model, where $\CW$ is the context window length. We  write it more transparently as 
$$ x^{(n)}_{t+1} \sim \hat{\belief}^{(n)}_t = \softmax\bigg(
W_{\text{out}}\, \AR_{\theta,t}\bigl(x^{(n)}_{t-\CW+1:t}\bigr) + b_{\text{out}}
\bigg),
$$
where $\AR_{\theta,t}$  defined in~\eqref{eq:causal_masking2} is a deterministic nonlinear state construction function  induced by the transformer. 
The structure of $\AR_{\theta,t}$ in terms of self attention is elaborated  in Appendix~\vref{app:rope}.

\subsection{Multimodal Extension: Vision-Language Models}  \index{vision-language model}
The autoregressive transformer framework and pre-training objective presented above  generalize seamlessly to multimodal vision-language models (VLMs), which now constitute many frontier systems including GPT-4o, Gemini-1.5/2.0, Claude-3.5, Grok-2 Vision, and open-source models such as LLaVA, PaliGemma, and Qwen-VL.

The modification is confined to the input embedding stage. An input image is encoded by a frozen or co-trained vision module (typically a Vision Transformer or CLIP-style encoder) into a sequence of visual features $v_1, \dots, v_L \in \mathbb{R}^{d_v}$.
These are projected into the LLM's embedding dimension:
\[
h_j^{\text{vision}} = W_{\text{proj}} v_j + b_{\text{proj}} \in \mathbb{R}^d.
\]
The visual embeddings are then concatenated (or interleaved) with text token embeddings and processed by the same causal transformer blocks.

Pre-training proceeds via the identical next-token prediction objective \(\mathcal{L}(\theta)\) over the joint sequence, often on large-scale interleaved image–text corpora (e.g., captions, documents with figures, web pages).
All subsequent mechanisms—causal masking, positional encodings (applied jointly to visual and text positions), attention, feed-forward networks, and the cross-entropy loss—remain unchanged.

Thus, the mathematical formulation developed in this part (from embeddings through autoregressive pre-training) applies uniformly to both unimodal text models and modern multimodal systems.
Alignment (Part 2) and generation (Part 3) similarly extend directly to the multimodal setting.

\partsection{Alignment Training.  RLHF, DPO, RSFT and RLVR}  \index{alignment}

Alignment training is an iterative process of post-training refinement that transform   raw pre-trained LLMs (often capable but not aligned with
human intent)  into helpful assistants like ChatGPT, Claude, Grok, 
and Llama--3.1--Chat. It  combines  Reinforcement Learning from Human Feedback (RLHF) to capture nuanced human preferences (together with Direct Preference Optimization (DPO) as an efficient implementation), Rejection Sampling Fine Tuning (RSFT) to stabilize the output distribution, and Reinforcement Learning from Verifiable Rewards  (RLVR) to enforce objective logical correctness.

\subsection{Reinforcement Learning from Human Feedback (RLHF)}  \index{reinforcement learning from human feedback (RLHF)}

After pretraining, the base LLM parameterized by $\theta$ produces the 
autoregressive next-token distribution 
$\hat\belief_t(x_{t+1}\mid x_{1:t})$ but is not aligned with human preferences.

We partition the parameters as $\theta = [\theta_{\text{trainable}}, \theta_{\text{frozen}}]$,
where $\theta_{\text{trainable}}$ denotes a restricted alignment specific set of parameters and  $\theta_{\text{frozen}}$
denotes the pretrained backbone LLM parameters. 
RLHF introduces a preference-optimization stage that updates only  $\theta_{\text{trainable}}$ while  keeping the pretrained 
backbone $\theta_{\text{frozen}}$ fixed or tightly constrained to preserve pretrained capabilities.
Mathematically, RLHF is implemented via policy-gradient optimization of a language-model policy against a learned reward model trained from human preferences, typically with KL regularization to a reference policy, rather than via Q-learning.

\paragraph{A. Reward Model.}
Human annotators (often paid contractors) rank multiple model-generated completions for each prompt $x$.  
From these pairwise or listwise comparisons, a separate reward model 
$R_\phi(x_{1:T})$ is trained (e.g., via Bradley--Terry or a binary classifier) 
to predict a scalar preference score for a full completion $x_{1:T}$. Here $\phi$ denotes the parameters of the reward model, trained from human preference comparisons to assign a scalar score $R_\phi(x_{1:T})$ to a full completion.

\paragraph{B. KL-Regularized Preference Optimization.}
Let $\pi_\theta(y \mid x)$ denote the probability of a full completion sequence $y$ given prompt $x$ under the autoregressive policy parameterized by $\theta$.
(This factors as the product of next-token distributions $\hat{\belief}_t$ via the chain rule; see~\eqref{eq:softmax_predict}.)
RLHF fine-tunes the trainable parameters $\theta_{\text{trainable}}$ to maximize expected reward while penalizing deviation from a fixed reference policy $\pi_{\text{ref}}$ (typically the pretrained or SFT model):
\[
\theta_{\text{trainable}}^\star
=\arg\max_{\vartheta}
\;
\E_{\substack{x\sim\mathcal{D}\\ y\sim\pi_\vartheta(\cdot\mid x)}}
\Big[
    R_\phi(y) 
    \;-\;
    \beta\, \KL\bigl(\pi_\vartheta(\cdot\mid x)\,
    \Vert\,\pi_{\text{ref}}(\cdot\mid x)\bigr)
\Big].
\]
Here $x\sim\mathcal{D}$ denotes sampling a prompt uniformly from the RLHF 
training dataset; i.e., the expectation is taken over the empirical 
distribution of human-labeled prompts. In practice, this KL penalty is implemented via token-level log-probability regularization rather than an exact sequence-level KL.

The KL term stabilizes training and preserves pretrained capabilities.  
The reward model parameters $\phi$ remain frozen during this stage.
This objective encourages the model to generate  outputs that the 
reward model judges as high quality, but  not to drift too far from the pretrained model; the KL penalty acts like an “elastic band’’ that keeps the updated policy close to the original capabilities while still moving it in the direction preferred by humans.

RLHF requires far fewer steps than pretraining; it is applied for 100M–10B tokens.

{\em Remark}. The “Which answer do you prefer?” prompts shown to end-users are not used for 
RLHF or to update the LLM's weights. They are collected only for product-level 
A/B testing, interface tuning, and large-scale analytics; the core LLM is 
trained exclusively from curated datasets and from preference labels produced 
by paid annotators during the offline alignment pipeline.

\paragraph{Direct Preference Optimization (DPO).}  \index{direct preference optimization (DPO)}
DPO replaces the RL policy-optimization stage of RLHF,
eliminating the need to explicitly train a reward model and run
policy-gradient RL. 
DPO  directly fits the policy to human preference comparisons.  
Suppose preference data consist of triples $(x, y^+, y^-)$ where, given prompt $x$, response $y^+$ is preferred to $y^-$.  
Let $\pi_\theta(y|x)$ denote the model and $\pi_{\mathrm{ref}}$ a fixed reference policy (typically the pre-trained model).  
DPO arises from the KL-regularized objective
\[ \max_{\pi}
\E_{\substack{x \sim \mathcal D\\ y \sim \pi(\cdot|x)}}
\big[
R(x,y)
- \beta \mathrm{KL}(\pi(\cdot|x)\|\pi_{\mathrm{ref}}(\cdot|x))
\big]  \]
At the optimum the reward can be written (up to a prompt-dependent constant)
as 
\[
R(x,y)
= \beta^{-1}\!\left(
\log \pi_\theta(y|x)
- \log \pi_{\mathrm{ref}}(y|x)
\right) + C(x).
\]
Assuming a Bradley--Terry preference model,
\[
P(y^+ \succ y^- \mid x)
= \sigma\!\big(\beta (R^+ - R^-)\big),
\]
one obtains the DPO loss
\[
\mathcal{L}_{\mathrm{DPO}}
= - \log \sigma\!\Big(
\beta \big[
\log \tfrac{\pi_\theta(y^+|x)}{\pi_\theta(y^-|x)}
-
\log \tfrac{\pi_{\mathrm{ref}}(y^+|x)}{\pi_{\mathrm{ref}}(y^-|x)}
\big]
\Big).
\]
This increases likelihood of preferred responses while decreasing
likelihood of rejected ones relative to the reference policy.

Thus preference alignment reduces to maximum-likelihood estimation
with a pairwise logistic loss, avoiding explicit RL optimization
while preserving the KL-regularized control interpretation.

{\em Remark}.
A key advantage of DPO is that the prompt-dependent constant $C(x)$ (which represents the log-partition function of the reward distribution) does not need to be estimated. When substituting the expression for $R(x,y)$ into the Bradley--Terry preference model, we observe:
\begin{align*}
P(y^+ \succ y^- \mid x) &= \sigma\Big(\beta \big[R(x, y^+) - R(x, y^-)\big]\Big) \\
&= \sigma\Big(\beta \big[ (\beta^{-1} \log \tfrac{\pi_\theta(y^+|x)}{\pi_{\text{ref}}(y^+|x)} + C(x)) - (\beta^{-1} \log \tfrac{\pi_\theta(y^-|x)}{\pi_{\text{ref}}(y^-|x)} + C(x)) \big]\Big) \\
&= \sigma\Big( \log \tfrac{\pi_\theta(y^+|x)}{\pi_{\text{ref}}(y^+|x)} - \log \tfrac{\pi_\theta(y^-|x)}{\pi_{\text{ref}}(y^-|x)} \Big).
\end{align*}
Because $C(x)$ is independent of the completion $y$, it cancels out exactly when taking the difference between the rewards of the preferred and rejected responses. This allows DPO to optimize the policy using only the log-likelihood ratios, avoiding the intractability of calculating the partition function in high-dimensional token spaces.

\subsection{Rejection Sampling Fine-Tuning (RSFT)}  \index{Rejection Sampling}

Rejection Sampling Fine-Tuning (RSFT), commonly referred to as the Reward ranked finite tuning (RAFT) or \textit{Best-of-$N$} method, is an offline, sample-based approach to alignment. It serves as a stable alternative or precursor to online policy gradient methods like PPO. RSFT reuses a reward model $R_\phi$ trained from human preference data to filter synthetic completions, allowing the model to learn from its own high-quality outputs via standard Supervised Fine-Tuning (SFT).

Formally, given a prompt $x$, the RSFT procedure proceeds as follows:
\begin{enumerate}
    \item Generate $N$ independent completions $\{y_1, \dots, y_N\} \sim \pi_\theta(\cdot \mid x)$, where $\pi_\theta$ is the current policy.
    \item Compute scalar rewards $\{R_\phi(y_i)\}_{i=1}^N$ using the frozen reward model $R_\phi: \mathcal{Y} \to \mathbb{R}$, which outputs higher values for completions preferred by humans.
    \item Select the top-scoring completion $y_{\text{best}} = \arg\max_i R_\phi(y_i)$ (or the top $k$ for $k \geq 1$).
    \item Accumulate the high-reward prompt-completion pair $(x, y_{\text{best}})$ into a dataset and perform SFT on $\pi_\theta$ using these pairs as training examples.
\end{enumerate}

Theoretical analysis shows that as $N \to \infty$, under standard assumptions of independent sampling and a temperature parameter $\tau > 0$, the distribution of selected completions $y_{\text{best}}$ approximates:
\[
\pi^*(y \mid x) \propto \pi_{\text{ref}}(y \mid x) \exp\!\big(R_\phi(y)/\tau\big).
\]
This is the analytic solution to the KL-regularized RLHF objective. Thus, RSFT allows the model to shift its distribution toward the optimal policy without the instability associated with online reinforcement learning.

\subsection{Reinforcement Learning with Verifiable Rewards (RLVR)}  \index{reinforcement learning with verifiable rewards (RLVR)}

While RSFT relies on a neural reward model to approximate human preference, Reinforcement Learning from Verifiable Rewards (RLVR) is employed for tasks with objective "ground truth" correctness, such as mathematics, symbolic logic, and computer programming. In RLVR, the ``verifiable'' aspect refers to the use of a deterministic programmatic verifier (e.g., a compiler, unit test, or math engine) instead of a stochastic reward model.

Unlike the offline nature of RSFT, RLVR typically utilizes online RL algorithms (such as PPO or GRPO) to update the policy $\pi_\theta$ based on binary or sparse rewards:
\begin{itemize}
    \item \textbf{Exploration:} The model generates multiple reasoning paths or code snippets.
    \item \textbf{Verification:} A rule-based verifier checks the final output (e.g., comparing a numeric answer to the ground truth).
    \item \textbf{Optimization:} The model is penalized for incorrect answers and reinforced for correct ones, encouraging the emergence of self-correction and Chain-of-Thought (CoT) reasoning.
\end{itemize}

RLVR is particularly effective for overcoming ``reward hacking,'' a phenomenon where models exploit the quirks of a neural reward model to achieve high scores without actually solving the task. By anchoring the reward in verifiable truth, RLVR pushes the model toward genuine procedural understanding. However, it remains complementary to RLHF, as it cannot be used for subjective tasks like creative writing or conversational etiquette where no absolute verifier exists.

\paragraph{Example. nanochat and RLVR}
A recent minimal open-source example that illustrates several alignment concepts discussed here is \nanochat{}~\cite{karpathy_nanochat} by Andrej Karpathy, a full-stack extension of the \nanoGPT{} repository for training ChatGPT-style models.
Beyond pretraining and supervised fine-tuning (SFT), \nanochat{} includes an optional reinforcement learning stage using a simplified GRPO algorithm directly on the GSM8K dataset.
Rewards are computed programmatically by extracting and verifying the final numeric answer against ground truth—exactly the verifiable reward signal described in RLVR—yielding improvements in mathematical reasoning without a learned reward model or human preference data.
The RL stage is commented out by default in the main \code{speedrun.sh} script and described as experimental/dataset-specific.
Key excerpt illustrating the core RL idea (simplified from the GRPO loop):

\begin{lstlisting}[language=Python, caption={Simplified GRPO-style reward computation in nanochat}]
# Sample multiple completions for each GSM8K prompt
completions = model.generate(prompts, num_samples=N)

# Extract final answer and compare to ground truth
rewards = []
for completion in completions:
    extracted_answer = extract_final_answer(completion)
    rewards.append(1.0 if extracted_answer == ground_truth else 0.0)

# Policy gradient update using token-level advantages derived from rewards
# (on-policy, mean-shifted, no KL penalty or clipping)
loss = - (log_probs * advantages).mean()
optimizer.step()
\end{lstlisting}

This provides a concrete, hackable illustration of RLVR at small scale, complementary to the more general RLHF pipeline.

\paragraph{Summary. A Unified View of Alignment Methods.}
Most modern alignment procedures can be interpreted as solving a
KL-regularized policy optimization problem of the form
\[
\max_{\pi}
\E_{\substack{x \sim \mathcal D\\ y \sim \pi(\cdot \mid x)}}
\Big[
R(x,y)
- \beta \,\KL\bigl(\pi(\cdot \mid x)\|
\pi_{\text{ref}}(\cdot \mid x)\bigr)
\Big],
\]
where $\pi_{\text{ref}}$ is a reference policy (typically the pretrained or
SFT model), and $R(x,y)$ is a task-dependent reward measuring response quality.
The KL term prevents catastrophic drift from pretrained capabilities while
allowing preference-driven adaptation.

Different alignment methods correspond to different choices of reward
construction and optimization strategy:

\begin{itemize}
\item \textbf{RLHF:} $R(x,y)$ is predicted by a learned reward model trained
from human preference comparisons, and the objective is optimized using
policy-gradient RL.

\item \textbf{DPO:} Human preference comparisons are used directly,
yielding a closed-form policy update that reduces optimization to
pairwise logistic maximum-likelihood training without explicit RL.

\item \textbf{RSFT:} Samples are drawn from the current policy and filtered
using the reward model; supervised fine-tuning on high-reward samples
approximates the same KL-regularized optimum offline.

\item \textbf{RLVR:} Rewards are produced by deterministic verifiers
(e.g., unit tests or symbolic solvers) rather than neural reward models,
and policy optimization proceeds via RL using these verifiable rewards.
\end{itemize}

Thus modern alignment pipelines differ mainly in how rewards are obtained
and how optimization is carried out, while sharing a common objective of
reward maximization subject to a KL constraint that preserves pretrained
capabilities.

\partsection{Generation (Inference)}

Having discussed pre-training (Part 1) and alignment training (Part 2), we now discuss  \emph{generation},
where the LLM uses its learned parameters to produce new text by sampling one 
token at a time from the probability distribution it has learned to predict.  
Generation is the trained LLM repeatedly answering the question it was trained 
to answer -- ‘Given everything I’ve seen so far, what’s the most likely next 
token?’ -- but now we get to choose how boldly or conservatively it answers by 
selecting a temperature parameter $\tau > 0$.

After training, the LLM  parameters (neural network weights) $\theta^\star$ are fixed.
The user-provided prompt consists of the initial tokens
$x_1,\ldots,x_{t_0}$. For $t \ge t_0$, subsequent tokens
$x_{t_0+1}, x_{t_0+2}, \ldots$ are generated autoregressively as follows:  the LLM  first computes the logits for the next token:
$ W^*_{\text{out}}\, s_t + b^*_{\text{out}}$,  where $s_t$ is the final hidden representation at position $t$ produced by the transformer.
With sampling temperature $\tau > 0$, the next-token $x_{t+1}$  is drawn  randomly from the $V$-dimensional probability vector $\belief_t^\tau$:
\begin{equation}
  \begin{split}
 x_{t+1}  \sim \belief_t^\tau &= \softmax\bigg(\frac{W^*_{\text{out}}\, F_{\theta^\star}\big(\emb[x_{1:t}]\big) \,\unit_t
           + b^*_{\text{out}} }{\tau}\bigg)   \in \Belief \\
& = \softmax\bigg(
\frac{W^*_{\text{out}}\, \AR_{\theta^*,t}\bigl(x_{t-\CW+1:t}\bigr) + b^*_{\text{out}}}{\tau}
  \bigg), \quad \tau > 0,
  \end{split}
\end{equation}
where $\AR_{\theta,t}(\cdot)$ denotes the transformer feature map (cf.~\eqref{eq:causal_masking2}),
and $\CW$ is the context window length.

This sampled  token  $x_{t+1}$ is appended to the sequence, yielding 
$
(x_1,\dots,x_t,x_{t+1})$, 
and the above  process is repeated.

A nice illustration of the above  process is the open source LLM nanoGPT; see \url{https://github.com/karpathy/nanoGPT}. In {\tt nanoGPT/model.py} the command
\begin{verbatim}  def generate(self, idx, max_new_tokens, temperature=1.0, top_k=None):
\end{verbatim}
implements the forward pass to get the logits, divides by temperature,  applies the softmax, and then samples the next token. 

\begin{lstlisting}[caption={Autoregressive token generation in \texttt{nanoGPT}},
                   label={lst:nanogpt_generate}]
for _ in range(max_new_tokens):

    # enforce finite context window x_{t-\Delta+1:t}
    idx_cond = idx if idx.size(1) <= self.config.block_size \
               else idx[:, -self.config.block_size:]

    # forward pass to compute hidden states and logits
    logits, _ = self(idx_cond)

    # extract final-step logits and apply temperature scaling)
    logits = logits[:, -1, :] / temperature

    # optional top-k truncation of the conditional distribution
    if top_k is not None:
        v, _ = torch.topk(logits, min(top_k, logits.size(-1)))
        logits[logits < v[:, [-1]]] = -float('Inf')

    # softmax to obtain \pi_t^\tau )
    probs = F.softmax(logits, dim=-1)

    # sample next token x_{t+1} \sim \pi_t^\tau
    idx_next = torch.multinomial(probs, num_samples=1)

    # append token and advance the context window
    idx = torch.cat((idx, idx_next), dim=1)

return idx
\end{lstlisting}

Generation continues until a stopping rule is triggered,
typically when the LLM emits a special end-of-sequence (EOS) token; a maximum-length
cap is used as a fallback. Modern LLM inference engines often employ dynamic stopping criteria beyond a fixed length limit -- such as halting early when the LLM assigns high probability to an EOS token or when additional heuristics (e.g., low logit variance or repeated token patterns) indicate that further generation would be redundant or low-quality---thereby accommodating highly variable response lengths efficiently.

{\em Remarks}: (i) 
The user prompt  is treated simply as the initial segment of the token sequence;
it has no special status once generation starts.
\\
(ii) At each generation step, the \textit{context} consists of the previously observed tokens within the context window, namely $x_{t-\CW+1:t}$. The user \textit{prompt} $x_1,\ldots, x_{t_0}$ is included in the context when $t \leq  t_0+ \CW-1$.
\\
(iii) Although the model is autoregressive in what it represents (each $x_{t+1}$ depends on past tokens),
a transformer computes hidden states for all positions in a context window in parallel.
Put simply: transformers are autoregressive in the distribution they model, but parallel in how they compute.
\\  (iv) 
{\em Temperature}.
$\tau=1$ recovers the exact distribution the LLM was trained on. 
Smaller temperature $\tau$ makes generation more deterministic (sharper softmax), 
while larger $\tau$ produces more diverse outputs. Indeed $\tau\to 0$ yields the greedy $\arg\max$ decoding, while $\tau\to \infty$ approaches a uniform distribution over the vocabulary.
 \\
(v) Although the dependence of $s_t$ on the input tokens is restricted to the most
recent $\CW$ tokens, this does not imply that  prompt information is forgotten during inference.
Rather, information from tokens $x_1,\ldots,x_{t-\CW}$ are represented implicitly in the hidden states of the tokens that
remain within the current context window.
Thus the transformer implements a finite-memory nonlinear autoregressive model
whose state acts as a learned sufficient statistic of the past.
\\
(vi)  {\em Catastrophic forgetting vs.\ context truncation.} In classical neural networks,  catastrophic forgetting means that the parameters forget old tasks after training on new tasks. In contrast, during LLMs inference, context truncation is due to loss of influence of earlier tokens once they fall outside the context window length $\CW$. This context truncation is a hard architectural constraint rather than a learning pathology, since the model parameters remain fixed during inference.

\subsection{Efficient Autoregressive Inference via KV Caching}

A naive implementation of the autoregressive model
in~\eqref{eq:AR_gen}--\eqref{eq:g_layer} is computationally prohibitive.
Since the hidden state $h_{\ell,t}$ depends on all preceding hidden states
$\{h_{\ell-1,s}\}_{s\le t}$, recomputing the full transformer state at each
generation step would result in $O(t^2)$ total computation for a sequence of
length $t$.

However, due to the causal masking property~\eqref{eq:causal_masking}, the hidden
states and attention projections of past tokens are invariant to future tokens.
This enables an efficient state-space realization of the LLM during inference,
known as \emph{key--value (KV) caching}.

\paragraph{State-Space Recursive Update.}
For each layer $\ell\in\{1,\dots,L\}$ and token position $s$, define the key and
value vectors
\[
{k}_{\ell,s} = N^{(\ell)}_\theta\, h_{\ell-1,s},
\qquad
{v}_{\ell,s} = L^{(\ell)}_\theta\, h_{\ell-1,s}.
\]
The KV cache at time $t$ is the collection
\[
\mathcal{S}_t
=
\bigcup_{\ell=1}^L
\bigl\{
({k}_{\ell,s},{v}_{\ell,s})
\bigr\}_{s=1}^t .
\]

When generating token $x_{t+1}$, the model does not recompute the entire history.
Instead:
\begin{itemize}
\item \emph{Current projections:} Only the most recent hidden state
$h_{\ell-1,t}$ is used to compute the query
${q}_{\ell,t} = M^{(\ell)}_\theta h_{\ell-1,t}$, along with
${k}_{\ell,t}$ and ${v}_{\ell,t}$.
\item \emph{Cache update:} The new $({k}_{\ell,t},{v}_{\ell,t})$
pair is appended to $\mathcal{S}_{t-1}$ to form $\mathcal{S}_t$.
\item \emph{Local attention:} The attention update uses the cached keys and
values:
\[
g^{(\ell)}_\theta(\cdot)
=
\phi\!\left(
\sum_{s=1}^{t}
\softmax_s\!\bigl(
\langle {q}_{\ell,t}, {k}_{\ell,s} \rangle
\bigr)\,
{v}_{\ell,s}
\right).
\]
\end{itemize}

Thus, $\mathcal{S}_t$ acts as a sufficient statistic (internal state) summarizing
the past sequence.

\paragraph{Computational and Memory Complexity.}
KV caching eliminates the need to recompute hidden states and projections for
past tokens, reducing the total inference cost from quadratic to linear in the
sequence length. The dominant per-token cost becomes the attention operation,
which scales as $O(t)$ due to the dot products with cached keys.

The memory requirement grows linearly with sequence length:
\[
\text{space complexity} = O(L\, d\, t),
\]
which can occupy several gigabytes of GPU memory for large models and long
contexts. As a result, modern LLM inference is typically limited not by raw
compute (FLOPS), but by memory bandwidth required to repeatedly access the KV
cache during attention computation.

The KV cache $\mathcal{S}_t$ effectively compresses the history of the conversation into a fixed-length representation for the current query $q_t$. While the transformer architecture is often praised for its "global attention," the practical implementation of generation reveals it to be a non-linear Markov process of order 1 over the state space of hidden representations.

\paragraph{Summary} Generation thus defines a deterministic controlled Markov process:
the state is $\mathcal{S}_t$, the action is the sampled token $x_{t+1}$, and the
state transition
$\mathcal{S}_{t+1}=\mathcal{T}_\theta(\mathcal{S}_t,x_{t+1})$
is deterministic.
Alignment methods such as RLHF and RLVR can therefore be interpreted as policy
optimization over this induced Markov decision process.

%%%%%%%%%%%%%%%%%%%%%%%%%%%%%%%%%%%%%%%%%%%%%%%

\subsection{Hallucination (inference-time fabrication).}   \index{hallucination}
During inference, an LLM samples tokens autoregressively from the predictive distribution
$\belief_t^\tau(\cdot\mid x_{1:t})$ in~\eqref{eq:AR_gen}.  We say the model \emph{hallucinates}
when the generated continuation contains a factual claim $c$ whose truth value is false
(unsupported by reliable  external sources), despite being assigned high probability by the model.

A simple way to formalize this is to introduce a task-dependent (possibly probabilistic)   verifier
$V(c)\in [0,1]$ indicating whether claim $c$ is correct under a reference source of truth
  (e.g., a database, retrieval system, or human fact-checker). Here $V(c)=1$ denotes a  fully supported claim while  $V(c)=0$ corresponds to a  clear falsehood.

  Let $c=c(y)$ be the claim(s)
extracted from a completion $y\sim \pi_{\theta^\star}(\cdot\mid x)$ and define the
\emph{hallucination indicator} $H(y)=1-V(c(y))$.  Then expected hallucination rate under prompt
distribution $\mathcal D$ is
\[
\mathrm{Hall}(\theta^\star)
\;\triangleq\;
\E_{x\sim\mathcal D}\;
\E_{y\sim \pi_{\theta^\star}(\cdot\mid x)}\big[\,H(y)\,\big].
\]
Equivalently, for a set of claims $\mathcal C(x)$ that might be asserted under prompt $x$,
hallucination corresponds to a mismatch between the model-implied probabilities
$P_{\theta^\star}(c\mid x)$ and the verifier-implied truth values:
high $P_{\theta^\star}(c\mid x)$ for claims with $V(c)=0$.
Temperature $\tau$ in~\eqref{eq:AR_gen} and decoding rules (e.g., top-$k$/nucleus sampling)
affect $\mathrm{Hall}(\theta^\star)$ by reshaping the sampling distribution, even though
the underlying conditional model $\pi_{\theta^\star}$ is unchanged.

{\em Dynamics of Hallucination}. \index{hallucination!dynamics}
Hallucination can  be interpreted dynamically in the autoregressive state evolution. Once a low-probability but coherent deviation occurs, subsequent predictions are conditioned on this perturbed context. Because transformer layers composed of residual connections and attention mechanisms do not impose strong corrective dynamics toward a unique semantic trajectory, such deviations may propagate across tokens rather than being rapidly corrected, leading generation to drift into coherent but factually incorrect continuations. This interpretation is consistent with analyses suggesting that hallucination arises from approximately neutral propagation of semantic deviations rather than systematic amplification or suppression.

Specifically, 
recall  the LLM generation process is  described by the AR model
\[ 
x_{t+1} \sim \belief_t^\tau = \softmax\!\left( \frac{W^*_{\text{out}} \, \AR_{\theta^*,t}\bigl(x_{t-\CW+1:t}\bigr) + b^*_{\text{out}}}{\tau} \right),
\]
with the hidden-state feature map defined as
\[
s_t = \AR_{\theta^*,t}\bigl(x_{t-\CW+1:t}\bigr) = F_{\theta^\star}\bigl(\emb[x_{t-\CW+1:t}]\bigr) \,\unit_t \in \mathbb{R}^d.
\]
Suppose at some step $t$ a low-probability but locally coherent token $x_t$ is sampled, such that the continuation deviates from a factually supported path (i.e., the implied claim $c(x_{1:t})$ receives low verifier score $V(c(x_{1:t})) \approx 0$).  
For all subsequent steps $k \ge 1$, the next-token distribution becomes
\[
\belief_{t+k}^\tau(\cdot \mid x_{1:t+k}) = \softmax\!\left( \frac{W^*_{\text{out}} \, \AR_{\theta^*,t+k}\bigl(x_{(t+k)-\CW+1:t+k}\bigr) + b^*_{\text{out}}}{\tau} \right),
\]
which is now conditioned on the perturbed prefix $x_{1:t+k}$ that already contains the initial deviation.

Because the transformer map $F_{\theta^\star}$ (and hence $\AR_{\theta^*,t}$) is composed of residual connections and attention mechanisms that are approximately \emph{neutral} with respect to small semantic perturbations in the input sequence 
(i.e., the map does not exhibit strong contractive dynamics toward a unique factual attractor nor pronounced instability that would rapidly amplify errors), 
such deviations tend to propagate rather than being quickly suppressed or corrected.  
Formally, the Jacobian of the hidden-state map with respect to earlier tokens often remains close to preserving small deviations across layers, allowing the trajectory $\{s_t\}_{t \ge t_0}$ in hidden space to drift away from regions corresponding to high-verifier-score continuations without encountering a strong restoring force.

Over many generation steps this results in coherent but factually unsupported long-range trajectories, even though each individual next-token prediction remains fluent under the learned conditional distribution $\pi_{\theta^\star}$.  
This dynamical picture is consistent with mechanistic interpretability results showing that residual streams in modern transformers often exhibit near-unit singular values in semantically relevant directions, leading to approximately neutral propagation of early factual errors rather than systematic amplification or rapid correction.

{\em Remarks}:
Alignment methods such as RLHF reduce hallucination indirectly by biasing the policy toward responses judged correct or helpful by human annotators, thereby lowering probability mass assigned to unsupported claims. When correctness is externally verifiable, RLVR instead provides direct reward signals that more explicitly suppress hallucinated outputs.

\subsection{Retrieval-Augmented Generation (RAG)}
\index{RAG}
\index{hallucination!mitigation}

Retrieval-Augmented Generation (RAG) is an inference-time technique that grounds the LLM's autoregressive generation in external, verifiable knowledge sources, thereby reducing the probability of drift into factually unsupported continuations.

Let the original user prompt be the token sequence $x_{1:t_0}$. In standard (vanilla) generation, the model conditions solely on this prefix via the hidden-state map
\[
s_t = \AR_{\theta^*,t}\bigl(x_{t-\CW+1:t}\bigr) = F_{\theta^\star}\bigl(\emb[x_{t-\CW+1:t}]\bigr) \,\unit_t \in \mathbb{R}^d,
\]
yielding the next-token distribution
\[
x_{t+1} \sim \belief_t^\tau = \softmax\!\left( \frac{W^*_{\text{out}} \, s_t + b^*_{\text{out}}}{\tau} \right).
\]

In RAG, an external retriever first queries a knowledge base (e.g., vector database of documents, Wikipedia, private corpora) using an embedding of the prompt (or a reformulated query). This returns a ranked list of relevant chunks $z_1, \dots, z_m$ (each $z_j$ a tokenized sequence of length $l_j$). These chunks are concatenated or interleaved with the original prompt to form an \emph{augmented context sequence}
\[
x'_{1:t'} = [z_1 : \dots : z_m] \concat [x_{1:t_0}],
\]
where $t' \le \CW$ (or longer in models with extended context windows), and the concatenation respects causal ordering (retrieved chunks typically prepended or placed before the user query).

Generation then proceeds autoregressively on the augmented sequence:
\begin{equation}
x_{t+1} \sim \belief_t^\tau = \softmax\!\left( \frac{W^*_{\text{out}} \, \AR_{\theta^*,t}\bigl(x'_{t-\CW+1:t}\bigr) + b^*_{\text{out}}}{\tau} \right),
\label{eq:rag_gen}
\end{equation}
with hidden states now
\[
s_t = \AR_{\theta^*,t}\bigl(x'_{t-\CW+1:t}\bigr) = F_{\theta^\star}\bigl(\emb[x'_{t-\CW+1:t}]\bigr) \,\unit_t \in \mathbb{R}^d.
\]

The key effect is that the hidden-state trajectory $\{s_t\}$ is conditioned on external evidence from the start, constraining the generative dynamics to remain closer to regions of the output space corresponding to high-verifier-score continuations. Formally, RAG reduces the expected hallucination rate
\[
\mathrm{Hall}(\theta^\star) = \E_{x\sim\mathcal{D}} \E_{y\sim \pi_{\theta^\star}(\cdot\mid x)} \bigl[ H(y) \bigr]
\]
by replacing the prompt distribution over ungrounded prefixes with one over grounded augmented prefixes $x'$, yielding
\[
\mathrm{Hall}_{\text{RAG}}(\theta^\star) = \E_{x\sim\mathcal{D}} \E_{z \sim \text{retriever}(\cdot\mid x)} \E_{y\sim \pi_{\theta^\star}(\cdot\mid x')} \bigl[ H(y) \bigr] \ll \mathrm{Hall}(\theta^\star)
\]
in knowledge-intensive tasks, where the retriever provides relevant factual support.

{\em Remarks}:
\begin{itemize}
  \item RAG is purely inference-time: it does not modify $\theta^\star$ or require retraining.
  \item The effectiveness depends on retriever quality (embedding model, indexing, reranking) and chunk relevance.
  \item In modern variants (2026), RAG often includes multi-hop retrieval, query reformulation, self-correction loops, or hybrid search (dense + sparse), further constraining drift.
  \item While very long context windows reduce the need for naive chunking, RAG remains essential for private/enterprise data, real-time information, and traceable provenance.
\end{itemize}

RAG complements alignment methods (RLHF, RLVR) by providing external grounding where internal knowledge is incomplete or outdated, making it one of the most deployed hallucination mitigations in production systems.

%%%%%%%%%%%%%%%%%%%%%%%%%%%%%%%%%%%%%%%%%%%%%%%%%

\subsection{Worked Example: One Step of LLM Generation}
Suppose the user prompt is:
\[
(x_1,x_2,x_3) = (\text{``The''},\,\text{``dog''},\,\text{``chased''})
\]

After computing the hidden state $s_3$, the LLM produces the following
logits for the next token over a tiny toy vocabulary
\[
\mathcal{V}=\{\text{cat},\text{ball},\text{car}\}:
\]
\[
z = W_{\text{out}} s_3 + b_{\text{out}}
= 
\begin{bmatrix}
2.0\\
1.0\\
0.1
\end{bmatrix}
\quad\text{corresponding to}\quad
\{\text{cat},\text{ball},\text{car}\}.
\]

{\bf Step 1: Convert logits to probabilities ($\tau=1$)}:
\[
\belief_3^1(i)  = \softmax(z/\tau)_i 
= \frac{e^{z_i}}{\sum_j e^{z_j}}
\]

Numerically,
$
e^{2.0}=7.39$, $ e^{1.0}=2.72$, $ e^{0.1}=1.11
$, $
\sum_j e^{z_j} = 11.22
$.
So the next-token probabilities are
\[
p(\text{cat})=0.659,\qquad
p(\text{ball})=0.242,\qquad
p(\text{car})=0.099.
\]

{\bf Step 2: Sample the next token}.
The LLM  draws a random sample from this distribution.  
The most likely outcome is
\[
x_4 = \text{``cat''}.
\]

The sequence now becomes
\[
(\text{``The''},\,\text{``dog''},\,\text{``chased''},\,\text{``cat''}).
\]

The same procedure is then repeated to generate $x_5,x_6,\dots$.

{\em Effect of the Temperature $\tau$}.
With a smaller temperature $\tau=0.5$, the distribution becomes sharper:
\[
\softmax(z/0.5) = \softmax([4.0,\,2.0,\,0.2]),
\]
which yields approximately
$
p(\text{cat}) \approx 0.90,
$
so the model behaves almost deterministically.

With a larger temperature $\tau=2$, the distribution becomes flatter:
\[
\softmax(z/2) = \softmax([1.0,\,0.5,\,0.05]),
\]
and so
$
p(\text{cat}) \approx 0.46$,
$p(\text{ball}) \approx 0.28$, 
$ p(\text{car}) \approx 0.26$.
Thus, larger values of $\tau$ produce more diverse  outputs.

%\subsection*{Summary. LLM as  High-Dimensional Nonlinear Autoregressive Model with Self-Attention}

\partsection{Extensions}

The topics discussed here are important in the design of LLMs but are best viewed as extensions of the basic ideas covered in Parts 1, 2 and 3. 

\subsection{Continual Learning (Selective Parameter Update Under Streaming Data)}

In continual learning, the new data $\mathcal{D}^{(k+1)}$ typically consists of 
fresh text corpora—new documents, updated domain-specific material, or 
synthetic text generated by the LLM itself—and the goal is to incorporate 
this new information without overwriting the knowledge encoded during 
pretraining.

A  challenge in adapting large pretrained LLMs is 
\emph{catastrophic forgetting}: 
na\"ively fine-tuning all parameters of the model on a new corpus 
$\mathcal{D}^{(k+1)}$ can  destroy the predictive performance that the model 
previously achieved on earlier corpora 
$\mathcal{D}^{(1)},\dots,\mathcal{D}^{(k)}$.
This occurs because gradient updates intended to fit the new data can alter 
the hidden representations $F_\theta(\Embed)$ in directions that overwrite
knowledge encoded during pretraining.

Continual learning addresses this problem by decomposing the parameter vector into
\[
\theta = (\theta_{\text{frozen}},\;\theta_{\text{trainable}}),
\]
where $\theta_{\text{frozen}}$ is the large pretrained backbone 
(embeddings, attention blocks, feedforward layers, LayerNorms, output matrix) 
that remains fixed, while $\theta_{\text{trainable}}$ is a much smaller set of 
parameters added specifically for ongoing adaptation 
(e.g., LoRA adapters, bias-only tuning, or small MLP expansions).

Given a new corpus $\mathcal{D}^{(k+1)}$, continual learning performs a selective 
update
\begin{equation}
\theta_{\text{trainable}}^{(k+1)}
=
\arg\min_{\vartheta}
\Big[
\mathcal{L}^{(k+1)}(\theta_{\text{frozen}}, \vartheta)
\;+\;
\lambda\,
\Omega\!\left(\vartheta,\;\theta_{\text{trainable}}^{(k)}\right)
\Big],  \label{eq:CL}
\end{equation}
where the regularizer $\Omega$ penalizes changes that would distort previously 
learned hidden representations $F_{\theta^{(k)}}(\Embed)$.
Thus the goal of continual learning is to 
\emph{incorporate new information without forgetting old information} by 
updating only the small subset of  parameters $\theta_{\text{trainable}}$
while keeping the pretrained backbone $\theta_{\text{frozen}}$ stable.

Continual learning for deployed LLMs is performed centrally by the
model provider at  update cycles (weeks to months), not continuously
after each user interaction. 
%, in order to control stability, safety, and
%catastrophic forgetting.

\paragraph{Alignment Training vs.\ Continual Learning.}
Although alignment training (Part~2) and continual learning  both modify 
only a small subset of parameters $(\theta_{\text{trainable}})$ while keeping the 
main backbone $(\theta_{\text{frozen}})$ fixed, they serve fundamentally 
different purposes.
Alignment training uses human preference data to adjust the model's behavior 
\emph{without changing its factual knowledge}: the goal is to make the LLM's 
responses helpful, honest, harmless, and consistent with human intent.
By contrast, continual learning addresses a different problem---\emph{updating 
the model with new information without forgetting old information}---typically 
under streaming or evolving data.  
Thus alignment shapes the model's \emph{preferences and behavior}, whereas 
continual learning updates the model's \emph{knowledge} while preventing 
catastrophic forgetting.

\paragraph{Catastrophic Forgetting.} \index{catastrophic forgetting}

Naïve sequential fine-tuning on new corpora causes catastrophic forgetting—
the rapid degradation of performance on earlier data—whereas continual learning 
introduces constraints and selective parameter updates to prevent this effect.
(We emphasize that this notion of catastrophic forgetting is distinct from inference-time context truncation, which arises from finite context windows and does not involve parameter updates.)

Indeed continual learning is designed to avoid catastrophic forgetting.
Catastrophic forgetting arises because gradient-based updates for a new corpus 
$\mathcal{D}^{(k+1)}$ typically move the parameter vector $\theta$ in 
directions that are not orthogonal to those responsible for performance on the 
earlier corpora $\mathcal{D}^{(1)},\dots,\mathcal{D}^{(k)}$.
Formally, the gradient of the new loss
\[
g^{(k+1)} = \nabla_\theta \mathcal{L}^{(k+1)}(\theta)
\]
generically has a nonzero projection onto the subspace spanned by the earlier 
task gradients
\[
\mathcal{S}^{(k)} 
= \mathrm{span}\!\left\{
\nabla_\theta \mathcal{L}^{(i)}(\theta) : i = 1,\dots,k
\right\}.
\]
Thus a single gradient step
\[
\theta \leftarrow \theta - \eta\, g^{(k+1)}
\]
modifies not only the features needed for the new corpus 
$\mathcal{D}^{(k+1)}$, but also the hidden representations 
$F_\theta(\Embed)$ used to solve the earlier corpora.
Since deep networks represent knowledge in a highly distributed fashion,
small parameter changes can produce large changes in internal activations.
As a result, the predictions for the earlier corpora can degrade rapidly---the 
phenomenon known as \emph{catastrophic forgetting}. 
This is not a form of geometric discounting or exponential forgetting; it is a 
consequence of the non-orthogonality of task gradients in high-dimensional 
parameter space.

{\em Remark.}
Neither \nanoGPT{}~\cite{karpathy_nanogpt} nor its successor \nanochat{}~\cite{karpathy_nanochat} implements continual learning techniques such as LoRA adapters or regularization to prevent catastrophic forgetting.
Both repositories support continued training on new corpora by loading a checkpoint and resuming optimization (using the same next-token objective), but this defaults to full-parameter fine-tuning---precisely the naïve approach that can lead to catastrophic forgetting on earlier data.
They serve as clean baselines illustrating the need for the selective parameter updates and regularization discussed above.

\subsection{Sycophancy as an Artifact of Alignment Training}  \index{sycophancy}
\index{alignment!sycophancy}
Alignment training (Part 2) shapes model behavior through preference optimization, but it introduces known side effects.
A prominent artifact is \emph{sycophancy}: the tendency of the aligned LLM to affirm, mirror, or agree with user statements, even when they are factually incorrect.
(Just as overfitting in supervised learning favors memorization over generalization, sycophancy in alignment training favors perceived agreeability over factual accuracy.)

Sycophancy emerges because human raters typically prefer responses that are polite, empathetic, non-confrontational, or validating—even on objective questions.
The resulting reward model $R_\phi(x_{1:T})$ systematically assigns higher scores to agreeable completions.
During optimization, the trainable parameters $\theta_{\text{trainable}}$ are updated to maximize the expected reward
\[
\mathbb{E}_{x_{1:T}\sim \pi_\theta}
\bigl[\, R_\phi(x_{1:T}) \,\bigr],
\]
which inadvertently shifts the aligned next-token distribution toward agreeable outputs:
\[
\hat{\belief}_t^{\text{aligned}}(\text{agreeable token} \mid x_{1:t})
\gg
\hat{\belief}_t^{\text{pretrained}}(\text{agreeable token} \mid x_{1:t}).
\]
Consequently, the model may produce flattering but misleading responses (e.g., entertaining conspiracy theories or avoiding direct correction of user errors).
Mitigating sycophancy through constitutional constraints, truthfulness rewards, or debiasing preference data—remains an active area of alignment research.

Finally,  \textit{hallucination differs from sycophancy}: Hallucination, discussed previously,  is primarily an \textit{inference-time dynamic phenomenon} arising when the model assigns high probability to unsupported or false claims. By contrast,  sycophancy is primarily an \textit{alignment-training time static artifact} in which preference optimization encourages agreement with user statements even when they are incorrect.  \index{hallucination!vs.\ sycophancy}
Put informally:
\begin{quote} Hallucination: model believes false things.
Sycophancy: model says what the user wants to hear.
\end{quote}

\subsection{Safety and Ethics}

Beyond sycophancy, alignment training does not fully resolve several persistent
safety and ethical challenges in LLMs. As discussed above, hallucination—generation of fluent but
factually incorrect or fabricated content—remains a major risk in
high-stakes domains such as medicine or law.
Bias amplification from training data can perpetuate societal inequities,  and may also be amplified rather than eliminated by
alignment procedures, leading to discriminatory outputs. 

From a security perspective, alignment remains vulnerable to adversarial
interaction: prompt injection and jailbreaking can circumvent behavioral
constraints despite safety training. More broadly, evaluating alignment is
difficult due to prompt sensitivity and reliance on subjective human judgments,
and recent work has raised concerns about goal misgeneralization in more agentic
LLM-based systems.
These issues indicate that current alignment methods, including RLHF, provide
only partial mitigation, motivating continued research into robustness,
debiasing, and more principled safety mechanisms.

\partsection{Prompting Strategies in Inference: In-Context Learning and Chain-of-Thought}

Part 3 discussed how a trained LLM generates text -- this is called the  generation or inference step. 
\textit{In-context learning} (ICL) and \textit{chain-of-thought} (CoT) prompting are both
\emph{inference-time} mechanisms: they alter how a fixed, trained LLM is
queried through the prompt and decoding process, without changing the model
parameters learned during pretraining or alignment.
Roughly speaking,  ICL changes the task via the prompt, while 
CoT changes the computation via the prompt.

\subsection{In-Context Learning (ICL).}   \index{in-context learning (ICL)}
In-context learning refers to the ability of a large language model to adapt to a
new task \emph{at inference time} by conditioning on a small number of examples
provided in the prompt, without any update of the model parameters.
Formally, for a fixed parameter vector $\theta$, the LLM generates the next token by sampling as  
\[
y \sim p(y \mid P, x;\theta),
\]
where $P$ is a prompt containing a few input--output demonstration pairs and
$x$ is a new query, and $y$ is the LLM's predicted output.  Although no optimization of $\theta$ occurs, the LLM
implicitly infers a task-specific input--output mapping from the prompt itself.
This phenomenon emerges only at large model scale and was first clearly
demonstrated in GPT--3, where zero-shot, one-shot, and few-shot performance
improves smoothly with parameter count. (Zero-shot is pure instruction without examples.)

\paragraph{ICL vs.\ Continual Learning.}
In-context learning and continual learning address different forms of adaptation.
ICL performs \emph{task adaptation through conditioning only}, with $\theta$
fixed and no risk of catastrophic forgetting, but is limited by the finite context length.
By contrast, continual learning performs \emph{parameter adaptation over time}
by updating $\theta$ sequentially on new datasets, see~\eqref{eq:CL}.
Thus ICL is ephemeral and memoryless, whereas continual learning enables
persistent long-term adaptation.

\subsection{2. Chain-of-Thought (CoT) Prompting.}  \index{chain-of-thought (CoT)} \index{prompting!CoT}
Chain-of-thought prompting is a technique for eliciting multi-step reasoning
from LLMs by explicitly encouraging the generation of intermediate reasoning
steps in the output. This often leads to significantly improved performance on
tasks that require structured computation, such as arithmetic, symbolic
reasoning, and multi-step logic.
Instead of directly producing a final answer, the LLM is prompted to generate
a sequence of logical or arithmetic steps leading to the conclusion.
For example, rather than answering “$17+25$,” the LLM is prompted to generate:
\[
17+25 = (10+20)+(7+5)=30+12=42.
\]
CoT emerges reliably only in sufficiently large LLMs (typically tens of
billions of parameters or more), where depth and attention enable iterative
composition across tokens.

Mathematically, CoT decomposes the output $y= (r_{1:K},a)$ where $r_{1:K}$ is the reasoning trajectory
 and $a$ denotes the final answer. Under CoT prompting, the LLM   samples from the joint distribution:
 $$ y = (r_{1:K},a) \sim p(a, r_{1:K} \mid P, x;\theta). $$
In comparison,  the standard LLM  samples from the marginal distribution as   $a \sim p(a \mid x;\theta)$ obtained  by marginalizing over the latent reasoning
variables $r_{1:K}$. Thus,  under CoT
prompting, the LLM is forced to sample an extended trajectory
$(r_{1:K}, a) \sim p(r_{1:K}, a \mid x;\theta)$ consisting of intermediate
reasoning steps followed by the final answer.

\medskip
\noindent
\textbf{Remark (Role of the Prompt).}
The user (or system designer) is responsible for crafting the prompt that
induces chain-of-thought behavior—for example by including phrases such as
“Let’s think step by step” (zero-shot CoT) or by providing few-shot examples
with explicit reasoning traces. The LLM then generates the step-by-step
reasoning and final answer during inference. Thus, CoT is a \emph{prompt-induced
computational mode}, not a separate training procedure. While larger modern
LLMs increasingly produce useful reasoning chains with minimal guidance, the
effectiveness of CoT still depends strongly on prompt structure.

\paragraph{Reasoning as an Emergent Behavior.}
Reasoning in LLMs is not explicitly programmed but emerges from large-scale
sequence modeling combined with architectural depth and attention-based
composition. Under CoT prompting, the transformer reuses its internal
representations across tokens to perform iterative computation.
This behavior is measured through benchmarks such as BIG-Bench and MMLU, where
LLMs exhibit planning, deduction, and compositional generalization far beyond
simple pattern matching.

\medskip
\noindent
\emph{In summary: ICL and CoT move the computation inside the prompt, not inside the weights.}

\subsection{Worked Examples: In-Context Learning vs.\ Chain-of-Thought}

We illustrate the difference between in-context learning (task adaptation)
and chain-of-thought prompting (computational guidance) using simple numerical
examples.

\paragraph{Example 1: In-Context Learning (Task Learned from the Prompt).}

Suppose the LLM has never been  trained on the task
“convert numbers to their squares,” but we provide a few examples in the prompt:
\begin{align*}
  \textbf{Prompt } P:
  \begin{array}{l}
\quad 2 \mapsto 4 \\
\quad 3 \mapsto 9 \\
    \quad 5 \mapsto 25
  \end{array}
\\
\textbf{Query } x: \quad  \text{Now compute: } 7 \mapsto ?
\end{align*}
The LLM then generates
\[
\boxed{7 \mapsto 49.}
\]

Here the LLM has inferred the \emph{task rule} “square the input” purely from
the few examples in the prompt.
No parameters are updated; the task is induced entirely through conditioning:
\[
p(y \mid P, x;\theta).
\]
Thus ICL  is \emph{task adaptation without parameter learning}.

\paragraph{Example 2: Chain-of-Thought (Same Task, but with Reasoning Traced).}

Now consider a prompt that explicitly requests intermediate reasoning:
\[
\begin{array}{l}
\textbf{Prompt:} \\
\quad \text{Let’s solve step by step. What is } 17+25?
\end{array}
\]
Instead of answering immediately, the LLM generates:
\[
17 + 25 = (10 + 20) + (7 + 5) = 30 + 12 = \boxed{42}.
\]

Here the \emph{task is already known} (addition), but the prompt forces the LLM
to expose the intermediate computational steps.
The reasoning is not an external algorithm—it is generated token by token by
the same autoregressive process used for all language output.

\paragraph{Example 3: ICL and CoT Used Together.}

Both mechanisms can be combined in a single prompt:
\[
\begin{array}{l}
\textbf{Prompt:} \\
\quad \text{Example: } f(2)=5, \; f(3)=7, \; f(4)=9.\\
\quad \text{Now compute } f(10) \text{ step by step.}
\end{array}
\]
The LLM may respond:
\[
f(x)=2x+1,\quad f(10)=2\cdot10+1=\boxed{21}.
\]

Here:
\begin{itemize}
  \item ICL infers the \emph{task rule} $f(x)=2x+1$ from examples,
  \item CoT forces the LLM to explicitly \emph{trace the computation}.
\end{itemize}

\paragraph{Conceptual Summary.} Both ICL and CoT adapt LLM behavior at inference time with fixed parameters:
\begin{itemize}
  \item In-context learning determines \emph{what task} the LLM performs.
  \item Chain-of-thought determines \emph{how the computation is carried out}.
\end{itemize}

\medskip
\noindent
\emph{In summary, ICL and CoT move the computation inside the prompt, not inside
the weights.}

Actually CoT  prompting exists in two regimes, both subsumed under ICL:
\begin{itemize}
\item  Few-shot CoT is conventional few-shot ICL with extended labels that contain reasoning chains.
\item  Zero-shot CoT (“Let’s think step by step”) appears demonstration-free but is still ICL: the instruction phrase leverages patterns seen during pre-training to condition the LLM into generating reasoning trajectories without explicit exemplars.
\end{itemize}
In both cases no parameters are updated; only the conditioning context changes.

\subsection*{Closing Remarks and Literature}
\addcontentsline{toc}{part}{Closing Remarks and Literature Review}

The modern LLM is best understood not as a static repository of facts, but as a dynamic system. While its parameters encapsulate the statistical structure of human knowledge, its behavior is a product of deliberate alignment constraints and inference-time conditioning. By formalizing these processes through the lenses of optimization and state-space theory, we move from empirical observation to principled engineering.

\paragraph{Literature} The number of papers in LLMs is growing exponentially and listing them in an elementary document such as this is futile.
\cite{Vas17} is the original self attention paper. \cite{SAL24} introduced  rotary position embedding. References for the additional topics introduced are:  RLHF \cite{OWJ22},  fine tuning \cite{DXG23},  RLVR \cite{Wen25}, DPO \cite{RSM23},
 CoT \cite{WWS22},  sycophancy \cite{STK23}, safety and ethics  \cite{PHS22}.

 \appendix

\section{Appendix. Multi-Head Self-Attention and Rotary Position Embeddings (RoPE)} \label{app:rope}

In~\eqref{eq:transformer_1}, the transformer neural network  was defined  abstractly as a nonlinear
map
\[
F_\theta:\;\Embed^{(n)} \longmapsto S^{(n)}\in\reals^{d\times T_n},
\qquad
s_t = S^{(n)}\unit_t,
\]
where the hidden state matrix  $s_t^{(n)}$ outputted by the transformer was 
used to produce the conditional distribution of the next token using~\eqref{eq:next_logit} and~\eqref{eq:softmax_predict}. 
We now specify the internal structure of $F_\theta$.

\paragraph{Layered Structure of $F_\theta$.}
The transformer $F_\theta$ is the composition of $L$ layers,
\[
F_\theta = \Layer^{(L)}\circ\cdots\circ\Layer^{(1)},
\qquad
H_\ell=\Layer^{(\ell)}(H_{\ell-1}),\quad H_0=\Embed^{(n)}.
\]
Here, $H_\ell\in \reals^{d\times T_n}$ denotes the  hidden-state matrix \textit{after} layer $\ell$. Note that $H_0 = \Embed^{(n)}$ are  the raw token embeddings.

Each layer consists of a self-attention block followed by a position-wise
feed-forward network:
\begin{equation}\label{eq:hidden_state_matrices}
\boxed{
\begin{aligned}
\tilde H_\ell
&=
H_{\ell-1}
+
\MultiHead_\ell\!\bigl(\LayerNorm_\ell(H_{\ell-1})\bigr)  \in \reals^{d\times T_n}, 
\\
H_\ell
&=
\tilde H_\ell
+
\FFN_\ell\!\bigl(\LayerNorm'_\ell(\tilde H_\ell)\bigr),
\qquad \ell=1,\dots,L.
\end{aligned}}
\end{equation}
Thus $F_\theta(\Embed^{(n)})=H_L$, namely the output of the $L$-th (final) layer of the transformer.

\subsubsection*{Multi-Head Self-Attention as a Bilinear Operator}  \index{self-attention}

 Self-attention lets the current token look back at all previous tokens and automatically learn — from their content — how much attention to pay to each one when predicting the next token,
regardless
of how far back they are. 
In "the dog that chased the cat …",
“dog” and “chased” is  more relevant than “the”. Self-attention computes this context-dependent importance weighting. Each   attention head  learns to focus on different linguistic patterns (one head might track subject-verb agreement, another might track which noun a pronoun refers to, etc.).
Specialization is not  designed or enforced — it is  an emergent property of random initialization + gradient descent on a rich prediction task. 
Multi-head attention runs many such mechanisms in parallel and combines them.

Each layer $\ell$ has $\nhead$ attention heads. 
Fix a layer $\ell$, attention head $k$, and let $d_h=d/\nhead$.
For each token $t$  and context window length $\CW$, attention is computed over the \textit{context window} $\I_t$ of previous tokens: \index{context window}
\[\I_t \defn \{t-\CW+1,\ldots,t\}, \text{ where } |\I_t| = \Delta ,
\]
where indices corresponding to tokens $t < 1$ are omitted.
%
%\[\I_t \defn \{\max(1,t-\CW+1),\ldots,t\}, \text{ where } |\I_t| = \min(t,\CW).
%\]
Then for each token $t$ and any earlier token $s \in I_t$, define the query, key and value vectors as 
\[
q_{\ell,t}^k = W_Q^{k,\ell} h_{\ell-1,t} \in \reals^{d_h},\quad
k_{\ell,s}^k = W_K^{k,\ell} h_{\ell-1,s} \in\reals^{d_h},\quad
v_{\ell,s}^k = W_V^{k,\ell} h_{\ell-1,s} \in\reals^{d_h},
\]
where  $W_Q^{k,\ell},W_K^{k,\ell},W_V^{k,\ell}\in\reals^{d_h\times d}$ are called the
query, key and value matrices, respectively. 

The (causal) attention scores are defined as \emph{bilinear functions of the hidden
states} $(h_{\ell-1,t},h_{\ell-1,s})$:  \index{self-attention!bilinear form}
\begin{equation}\label{eq:attention_scores}
\boxed{
a_{\theta,\ell}^k(t,s)
=
\frac{(q_{\ell,t}^k)^\top (R_t^\top R_s)\,k_{\ell,s}^k}{\sqrt{d_h}} = \frac{
h_{\ell-1,t}^\top
\Big(
(W_Q^{k,\ell})^\top R_t^\top R_s\, W_K^{k,\ell}
\Big)
h_{\ell-1,s}
}{\sqrt{d_h}},
\qquad s \in \I_t,
}
\end{equation}
where the matrices $R_u \in \reals^{d_h \times d_h}$ implement rotary position embeddings (defined below).  The attention scores are unconstrained real-valued scalars; nonnegativity and normalization arise after applying the softmax, yielding the attention weights that form a $\CW$-dimensional probability vector:
\begin{equation}
  \label{eq:causal_attention}
w_{\theta,\ell}^k(t,\I_t)=\softmax\!\bigl(a_{\theta,\ell}^k(t,\I_t)\bigr)\in\Simplex(\CW).
\end{equation}
Finally  the $k$-th attention  head output at token $t$ in layer $\ell$ is 
\begin{equation}
  \label{eq:boxedattn}
\boxed{\Attn_k^\ell(h_{\ell-1,t})
=
\sum_{s \in \I_t} w_{\theta,\ell}^k(t,s)\,v_{\ell,s}^k
=
\sum_{s \in \I_t}
\softmax_s\!\left( 
\frac{ \big[ R_t\, W_Q^{k,\ell} h_{\ell-1,t}\big]^\p\,
 R_s
W_K^{k,\ell} h_{\ell-1,s}
}{\sqrt{d_h}}
\right)
\;
\bigl(W_V^{k,\ell} h_{\ell-1,s}\bigr).}
\end{equation}
Note that $\Attn_k^\ell:
\reals^d \to \reals^{d_h}$ where  $d_h={d}/{\nhead}$.
Also, for fixed token 
$t$, and any vector $z(t,\I_t)  \in \reals^{\CW}$,  $\softmax_s(z(t,s))$ is the $s$-th component of
 probability vector $\softmax(z(t,\I_t) )\in \Simplex(\CW)$.

Concatenating all $\nhead$ heads and applying an output projection matrix $W_0^\ell$ gives
\[
\MultiHead_\ell(X) 
=
\big[\Attn_1^\ell(X)\;\|\;\cdots\;\|\;\Attn_{\nhead}^\ell(X)\big]\,W_O^\ell \in \reals^{d \times T_n}, 
\qquad \text{ where } W_O^\ell\in\reals^{d\times d}.
\]

{\em Remarks:} 1.  The hidden state 
$h$ enters the attention mechanism in~\eqref{eq:boxedattn}  three times: as query, key, and value.
Thus, each attention head implements a \emph{bilinear similarity kernel}
followed by a \emph{linear state aggregation}.

2. The attention score $a_{\theta,\ell}^k(t,s)$ is best viewed as a learned similarity score: how much does the current token $t$ match the earlier token $s$. The larger the score, the more token $s$ will contribute to predicting the next token. 
If the current context is  ‘dog’ and earlier token was ‘dog’ (or ‘cat’, ‘chased’), the embedding makes their projected query and key vectors point in the same direction, so their dot product explodes, and the LLM indicates ‘This earlier token is highly  relevant right now!’

3. {\em  Matrix Form of Attention (Per-Head  Decomposition).}
Multi-head attention is obtained by applying the following construction
for each head $k=1,\ldots,\nhead$ and then concatenating the
results followed by the output projection $W_O^\ell$:
Stack the rotated query, key, and value vectors over all tokens
\[
Q_k \;\defn\; 
\begin{bmatrix}
R_1 q_{\ell,1}^k & \cdots & R_T q_{\ell,T}^k
\end{bmatrix}
\in\reals^{d_h\times T},
\qquad
K_k \;\defn\;
\begin{bmatrix}
R_1 k_{\ell,1}^k & \cdots & R_T k_{\ell,T}^k
\end{bmatrix}
\in\reals^{d_h\times T},
\]
\[
V_k \;\defn\;
\begin{bmatrix}
v_{\ell,1}^k & \cdots & v_{\ell,T}^k
\end{bmatrix}
\in\reals^{d_h\times T},
\qquad d_h \defn d/\nhead .
\]
Then the attention output for head $k$ over the entire sequence can be written
compactly as: \index{self-attention!matrix form}
\[
\boxed{ \text{Given } H_{\ell-1} \in \reals^{d \times T_n}, \quad 
\Attn_k^\ell(H_{\ell-1})
=
V_k\,
\softmax\!\Biggl(
\frac{K_k^\top Q_k}{\sqrt{d_h}}
\Biggr)
\;\in\;\reals^{d_h\times T},
}
\]
where the softmax is applied \emph{column-wise} so that the $t$-th column yields
the attention weights $w_{\theta,\ell}^k(t,\cdot)$.
Causality is enforced by setting the upper triangle of the matrix $K_k^\top Q_k$ to $-\infty$ before $\softmax$.

4.  Instead of~\eqref{eq:causal_attention}, for \emph{non-causal} bidirectional self-attention (e.g., in BERT-type encoders),
the softmax normalization is taken over all tokens in the sequence, i.e., $w^k_{\theta,\ell} (t,\cdot)$ is a $T$-dimensional probability vector:
$$ w^k_{\theta,\ell} (t,\cdot)= \softmax\bigl(a_{\theta,\ell}^k(t,1:T)\bigr) \in \Simplex(T). $$

5. In \nanoGPT{}, multi-head causal self-attention is implemented in the \texttt{CausalSelfAttention} module, with query/key/value projections, scaled dot-product attention, and a causal mask to enforce autoregression:

\begin{lstlisting}[language=Python, caption={Key lines from CausalSelfAttention in nanoGPT}]
# Q, K, V projections
q, k, v = self.c_attn(x).split(self.n_embd, dim=2)

# Causal mask (lower triangular)
mask = torch.tril(torch.ones(T, T, device=x.device)).view(1, 1, T, T)
att = (q @ k.transpose(-2, -1)) * (1.0 / math.sqrt(k.size(-1)))
att = att.masked_fill(mask[:,:,:T,:T] == 0, float('-inf'))
att = F.softmax(att, dim=-1)
y = att @ v  # (B, nh, T, T) x (B, nh, T, hs) -> (B, nh, T, hs)
\end{lstlisting}

%The main takeaway is: 
%the score matrix $K_k^\top Q_k$ is \emph{bilinear in the hidden states}
%because $Q_k$ and $K_k$ are linear functions of $H_{\ell-1}$ and include the
%relative-position operator $R_t^\top R_s$ via RoPE, while the output is a
%linear aggregation of values $V_k$ outside the softmax. Operationally, the hidden state 
%$h$ enters the attention mechanism three times: as query, key, and value.
%Thus each attention head implements a \emph{bilinear similarity kernel}
%followed by a \emph{linear state aggregation}.

\subsubsection*{Rotary Position Embeddings (RoPE)} Because transformers contain no recurrence or convolution, self-attention is  \index{Rotary Position Embeddings (RoPE)} \index{transformer architecture!RoPE}  \index{positional encoding!RoPE}
permutation-equivariant with respect to token order unless positional information is injected; positional
encodings allow the model to represent word order, for example, the 1st token from the 10th.
Positional information is provided by \textit{Rotary Position Embeddings} (RoPE), which make attention scores depend on the relative distance between tokens without  adding positional vectors to the input.  RoPE is used in recent versions of GPT.

RoPE encodes position by rotating query and key vectors within each head.
For $u\in\mathbb{N}$,
\[
R_u=\blockdiag\!\left(
\begin{pmatrix}\cos(u\theta_j)&-\sin(u\theta_j)\\
\sin(u\theta_j)&\cos(u\theta_j)\end{pmatrix}, j=1,\ldots, d_h/2
\right)\in\reals^{d_h\times d_h},
\qquad
\theta_j=10000^{-2(j-1)/d_h}.
\]
The identity $R_t^\top R_s = R_{\,s-t}$ shows that RoPE makes attention depend on the
\emph{relative offset} $t-s$, rather than absolute positions.
Causality is enforced by restricting $s\le t$.

{\em Remarks}. \index{positional encoding!fixed sinusoidal}
 (i) Early transformer architectures employed fixed sinusoidal (cosine--sine) absolute
positional encodings:  In Step 3, $e^{(n)}_t$ was replaced with
  $${e}^{(n)}_t + {p}_t$$ where  
  position information vector $p_t \in \reals^d$ had elements
  \[ 
p_t(2j) = \sin\!\left(t / 10000^{\,2j/d}\right), \qquad
p_t(2j+1) = \cos\!\left(t / 10000^{\,2j/d}\right), \quad j=0,1,\ldots, d/2-1.
\]
Subsequent GPT-style LLMs instead adopted learned absolute positional embeddings.
More recent architectures often employ relative positional encodings, such as
rotary positional embeddings (RoPE), which incorporate position information
directly into the attention mechanism at each transformer layer.

(ii) Word-order sensitivity (``dog ate man'' vs.\ ``man ate dog'') arises because of
the causal mask: each token only attends leftward, so the context differs.

(iii) Relative positional encodings are crucial for long-range dependencies and length generalization.
``The key that opens the door [50 tokens later]  was on the table''.

(iv) In nanoGPT, instead of a fixed sinusoid,  the $d$ dimensional positional column vector $p_t= P[t]$, i.e., $t$-th column of matrix $P \in {\reals^d \times \CW}$, where $P$ is a learned matrix and $\CW$ is the context window length.
The implementation in \nanoGPT{}'s \texttt{model.py} defines the positional embedding table as
\begin{lstlisting}[language=Python, caption={Positional embeddings in nanoGPT (from \texttt{model.py})}]
# Definition in GPT.__init__
self.transformer.wpe = nn.Embedding(config.block_size, config.n_embd)

# Usage in GPT.forward
tok_emb = self.transformer.wte(idx)  # token embeddings of shape (b, t, n_embd)
pos_emb = self.transformer.wpe(torch.arange(t, device=idx.device))  # position embeddings of shape
x = self.transformer.drop(tok_emb + pos_emb)
\end{lstlisting}

\paragraph{Layer Normalization and FFN} \index{transformer architecture!Layer Normalization}
Recall the generation of the hidden state matrices in~\eqref{eq:hidden_state_matrices} 
that comprises  head concatenation,  LayerNorms, and  FFN.

    Layer normalization is a  per-token mean/variance normalization + scale/shift -
    it allows us to
 stack hundreds of  transformer layers without the training going unstable.
    For any hidden vector $z \in \reals^d$:
\[
\LayerNorm(z)
\;=\;
\gamma \;\odot\;
\frac{z  - \mu}{\sqrt{\sigma^2 + \varepsilon}}
\;+\;
\beta,
\]
\begin{itemize}
  \item $\mu     = \frac{1}{d} \sum_{i=1}^d z_i$ \quad (mean over the $d$ coordinates of this  token)
  \item $\sigma^2 = \frac{1}{d} \sum_{i=1}^d (z_i - \mu)^2$ \quad (variance over the $d$ coordinates of this  token)
  \item $\gamma, \beta \in \mathbb{R}^d$ are learned parameters (different for each LayerNorm instance)
  \item $\varepsilon \approx 10^{-5}$ is a tiny constant for numerical stability
  \item $\odot$ denotes element-wise multiplication.
\end{itemize}

Finally,  the position-wise feed-forward network inside each layer is
\begin{equation}
  \label{eq:FFN}
    \operatorname{FFN}(x)
    \;=\;
    W_2 \, \sigma\!\bigl( W x + b_1 \bigr) + b_2
    \qquad (\sigma = \operatorname{GeLU}\;\text{or}\;\operatorname{SwiGLU})
  \end{equation}
  % \[ \text{ with components } 
%  \operatorname{SwiGLU}(x_i) = \frac{x_i^2}{1 + \exp(-x_i)} , \quad
%\operatorname{GeLU}(x_i) = x_i \Phi(x_i).
%\]
  
In \nanoGPT{}, layer normalization uses PyTorch's standard \texttt{nn.LayerNorm} (with the same per-token statistics and learned $\gamma, \beta$ parameters; $\varepsilon = 10^{-5}$ by default) and is applied twice per transformer block---after multi-head attention and after the FFN.
The position-wise feed-forward network employs two linear layers with a GeLU activation and a final projection, as implemented in the \texttt{MLP} module:

\begin{lstlisting}[language=Python, caption={LayerNorm and FFN in nanoGPT (from \texttt{model.py})}]
class FeedForward(nn.Module):  # also called MLP in the code
    def __init__(self, config):
        super().__init__()
        self.c_fc    = nn.Linear(config.n_embd, 4 * config.n_embd)  # W_1, b_1 implicit
        self.gelu    = nn.GELU()
        self.c_proj  = nn.Linear(4 * config.n_embd, config.n_embd)  # W_2, b_2 implicit

    def forward(self, x):
        return self.c_proj(self.gelu(self.c_fc(x)))

# In each Block (transformer layer)
x = x + self.attn(self.ln_1(x))      # post-LayerNorm after attention
x = x + self.ffwd(self.ln_2(x))     # post-LayerNorm after FFN
\end{lstlisting}

This matches the GeLU-based formulation in~\eqref{eq:FFN} and follows GPT-2's exact architecture.

%The above  formulation summarizes the  architecture of modern LLMs
%(Llama, Mistral, Gemma 2, PaLM).

\subsection*{Worked Example: Single-Head Self-Attention (Toy Example)}

 Suppose $d=2$ (embedding dimension) and the current sequence has $T=3$ tokens with hidden states
\[
h_1 =
\begin{bmatrix}1\\0\end{bmatrix},
\qquad
h_2 =
\begin{bmatrix}0\\1\end{bmatrix},
\qquad
h_3 =
\begin{bmatrix}1\\1\end{bmatrix}.
\]

Let the (single-head) query, key, and value matrices be
\[
 W_Q =
\begin{bmatrix}
2 & 0\\
0 & 1
\end{bmatrix}, W_K = W_V =
\begin{bmatrix}
1 & 0\\
0 & 1
\end{bmatrix}; \quad \text{ then } q_t = W_Q h_t \text{ (query) } , \quad
k_s = W_K h_s \text{ (key) }, \quad
v_s = W_V h_s \text{ (value) }.
\]

Thus
\[
q_3 = W_Q h_3 = \begin{bmatrix}2\\1\end{bmatrix},\qquad
k_1 = v_1=\begin{bmatrix}1\\0\end{bmatrix},\;
k_2 = v_2=\begin{bmatrix}0\\1\end{bmatrix},\;
k_3 = v_3= \begin{bmatrix}1\\1\end{bmatrix}.
\]

\paragraph{Step 1: Attention Scores (Dot Products)}
For causal attention at time $t=3$, only tokens $s \le 3$ are visible.  
The unnormalized attention scores are
\[
a(3,1) = q_3^\top k_1 = [2\;1]\!\begin{bmatrix}1\\0\end{bmatrix} = 2,
\quad
a(3,2) = q_3^\top k_2 = [2\;1]\!\begin{bmatrix}0\\1\end{bmatrix} = 1,
\quad
a(3,3) = q_3^\top k_3 = [2\;1]\!\begin{bmatrix}1\\1\end{bmatrix} = 3.
\]
So the attention  score vector is
\[
a_3 = \begin{bmatrix}2\\1\\3\end{bmatrix}.
\]
So token 3 finds token 3 most relevant, token 1 next, and token 2 least relevant.

\paragraph{Step 2: Softmax Normalization}
Applying softmax gives the attention weights:
\[
w(3,s) = \frac{e^{a(3,s)}}{\sum_{j=1}^3 e^{a(3,j)}} \implies w(3,1)=0.245,\qquad
w(3,2)=0.090,\qquad
w(3,3)=0.665.
\]

\paragraph{Step 3: Weighted Sum of Values}
The updated representation at position $t=3$ is
\[
\tilde h_3
= \sum_{s=1}^3 w(3,s)\, v_s
= 0.245
\begin{bmatrix}1\\0\end{bmatrix}
+ 0.090
\begin{bmatrix}0\\1\end{bmatrix}
+ 0.665
\begin{bmatrix}1\\1\end{bmatrix} = \begin{bmatrix}
0.910\\
0.755
\end{bmatrix}.
\]
The updated hidden state after this attention sublayer can be written as
\[
h_{\ell,3} = G_\theta(\tilde h_3),
\]
where $G_\theta$ denotes the remaining components of the transformer layer $l$.

\section{Glossary of Key Terms}

\begin{description}

\item[Alignment training {\normalfont[Alignment]}]
Post-training procedures that adapt a pretrained LLM to produce outputs aligned with human preferences or task objectives.

\item[Attention {\normalfont[Training/Inference]}]
Mechanism allowing tokens to interact by computing similarity-based weighted combinations of representations.

\item[Autoregressive model {\normalfont[Training/Inference]}]
Sequence model generating each token conditioned on previously generated tokens.

\item[Chain-of-Thought (CoT) {\normalfont[Inference]}]
Reasoning style where intermediate reasoning steps are generated before the final answer.

\item[Context {\normalfont[Inference]}]
Tokens visible to the model when predicting the next token.

\item[Context window {\normalfont[Training/Inference]}]
Maximum number of tokens the model conditions on at each generation step.

\item[DPO (Direct Preference Optimization) {\normalfont[Alignment]}]
Preference-alignment method reducing policy optimization to pairwise logistic likelihood training without explicit RL.

\item[Embedding {\normalfont[Training/Inference]}]
Mapping from discrete tokens to continuous vector representations used as neural network inputs.

\item[Hallucination {\normalfont[Inference]}]
Generation of fluent but factually incorrect or fabricated information not supported by training data or context.

\item[Inference {\normalfont[Inference]}]
Process of generating text from a trained model without updating parameters.

\item[In-context learning (ICL) {\normalfont[Inference]}]
Ability of an LLM to adapt to tasks using prompt examples without parameter updates.

\item[KV cache {\normalfont[Inference]}]
Stored key–value attention states enabling efficient autoregressive generation.

\item[Multi-head self-attention {\normalfont[Training/Inference]}]
Extension of self-attention using multiple parallel attention heads to capture diverse token relationships.

\item[Prompt {\normalfont[Inference]}]
User-provided input text that initializes generation. The prompt is a subset of  the context, provided it remains within the context window.

\item[RAG (Retrieval-Augmented Generation) {\normalfont[Inference]}]
  Inference-time technique that augments the prompt with externally retrieved relevant documents or chunks to ground generation in verifiable knowledge, reducing hallucination risk without modifying model parameters.
  
\item[RLHF (Reinforcement Learning from Human Feedback) {\normalfont[Alignment]}]
Alignment method using a learned reward model trained from human preference comparisons.

\item[RLVR (Reinforcement Learning from Verifiable Rewards) {\normalfont[Alignment]}]
Alignment using programmatically verifiable correctness signals rather than neural reward models.

\item[RoPE (Rotary Positional Encoding) {\normalfont[Training/Inference]}]
Relative positional encoding method that rotates query and key vectors to encode token position within attention computations.

\item[RSFT (Rejection Sampling Fine-Tuning) {\normalfont[Alignment]}]
Alignment procedure selecting high-reward outputs and retraining via supervised fine-tuning.

\item[Self-attention {\normalfont[Training/Inference]}]
Attention mechanism where tokens attend to other tokens in the same sequence.

\item[Softmax {\normalfont[Training/Inference]}]
Normalization function converting logits into a probability distribution.

\item[Sycophancy {\normalfont[Alignment]}]
Model behavior that excessively agrees with or flatters user input even when it conflicts with factual correctness.

\item[Temperature {\normalfont[Inference]}]
Sampling parameter controlling randomness by scaling logits before softmax.

\item[Token {\normalfont[Training/Inference]}]
Discrete unit of text processed by the model, typically a subword segment.

\item[Tokenization {\normalfont[Training]}]
Procedure converting raw text into discrete tokens used during model training and inference.

\end{description}

\bibliographystyle{IEEEtran}
\bibliography{$HOME/texstuff/styles/bib/vkm}

\printindex

\end{document}